\documentclass[10pt,journal,compsoc]{IEEEtran}

\usepackage{microtype}
\usepackage{graphicx}
\usepackage{subfigure}
\usepackage{booktabs}
\usepackage{hyperref}

\usepackage{amsmath}
\usepackage{amssymb}
\usepackage{esint}
\usepackage{times}
\usepackage{epsfig}
\usepackage{epstopdf}
\usepackage{caption}

\usepackage[utf8]{inputenc} % allow utf-8 input
\usepackage[T1]{fontenc}    % use 8-bit T1 fonts
\usepackage{hyperref}       % hyperlinks
\usepackage{url}            % simple URL typesetting
\usepackage{booktabs}       % professional-quality tables
\usepackage{amsfonts}       % blackboard math symbols
\usepackage{nicefrac}       % compact symbols for 1/2, etc.
\usepackage{microtype}      % microtypography
\usepackage{xcolor}
\usepackage{pifont}% http://ctan.org/pkg/pifont
\usepackage{wrapfig,lipsum,booktabs}
\usepackage[makeroom]{cancel}

\usepackage{xspace}

\newcommand{\myparagraph}[1]{\vspace{.5em}\noindent\textbf{#1}}
\newcommand{\bx}{\mathbf{x}\xspace}
\newcommand{\bO}{\mathbf{O}\xspace}
\newcommand{\bw}{\mathbf{w}\xspace}
\newcommand{\by}{\mathbf{y}\xspace}
\newcommand{\bY}{\mathbf{Y}\xspace}
\newcommand{\bpi}{\boldsymbol{\pi}\xspace}
\newcommand{\bPi}{\boldsymbol{\Pi}\xspace}
\newcommand{\balpha}{\boldsymbol{\alpha}\xspace}

\newcommand{\calY}{\mathcal{Y}\xspace}
\newcommand{\calL}{\mathcal{L}\xspace}
\newcommand{\D}{\mathcal{D}\xspace}

\newcommand{\iid}{\emph{i.i.d.}\xspace}
\newcommand{\ie}{\emph{i.e.}\xspace}
\newcommand{\eg}{\emph{e.g.}\xspace}
\newcommand{\cf}{\emph{cf.}\xspace}
\newcommand{\wrt}{\emph{w.r.t.}\xspace}
\newcommand{\Eq}{Eq.\xspace}
\newcommand{\Fig}{Fig.\xspace}
\newcommand{\Eqs}{Eqs.\xspace}
\newcommand{\Sec}{Sec.\xspace}
\newcommand{\Tab}{Tab.\xspace}

\newcommand{\revised}[1]{\textcolor{black}{#1}}

\newcommand{\xmark}{\ding{55}}%

\newcommand{\appropto}{\mathrel{\vcenter{
			\offinterlineskip\halign{\hfil$##$\cr
				\propto\cr\noalign{\kern2pt}\sim\cr\noalign{\kern-2pt}}}}}

\ifCLASSOPTIONcompsoc
  \usepackage[nocompress]{cite}
\else
  \usepackage{cite}
\fi

\ifCLASSINFOpdf
  
\else
  
\fi

\begin{document}

\title{Learn to Predict Sets Using Feed-Forward Neural Networks}

\author{Hamid Rezatofighi, Tianyu Zhu, Roman Kaskman, Farbod T. Motlagh, Javen Qinfeng Shi, \\Anton Milan, Daniel Cremers, Laura Leal-Taix{\'e}, Ian Reid % <-this % stops a space
\IEEEcompsocitemizethanks{\IEEEcompsocthanksitem Hamid Rezatofighi is with the Department of Data Science and AI, Faculty of Information Technology, Monash University, Melbourne, Australia.\protect\\
E-mail: hamid.rezatofighi@monash.edu
\IEEEcompsocthanksitem Tianyu Zhu is with Department of Electrical and Computer Systems Engineering, Monash University, Melbourne, Australia.
\IEEEcompsocthanksitem Farbod T. Motlagh, Javen Qinfeng Shi, and Ian Reid are with the School of Computer Science, The university of Adelaide, Australia.
\IEEEcompsocthanksitem Anton Milan is with Amazon. This work was done prior to joining Amazon.
\IEEEcompsocthanksitem Roman Kaskman, Daniel Cremers, and Laura Leal-Taix{\'e}, are with Technical University of Munich, Germany.}}% <-this % stops a space
% Manuscript received April 19, 2005; revised August 26, 2015.}}


\markboth{}%
{Shell \MakeLowercase{\textit{et al.}}: Bare Advanced Demo of IEEEtran.cls for IEEE Computer Society Journals}



\IEEEtitleabstractindextext{%
\begin{abstract}
This paper addresses the task of set prediction using
deep feed-forward neural networks. A set is a collection of elements
which is invariant under permutation and the size of
a set is not fixed in advance. Many real-world problems, such as image tagging and object
detection, have outputs that are naturally expressed as sets of entities.
This creates a challenge for traditional deep neural networks which naturally deal with structured outputs such as vectors, matrices or tensors. 
We present a novel approach for learning to predict sets with unknown permutation and cardinality using deep neural networks. In our formulation we define a likelihood for a set distribution represented by a) two discrete distributions defining the set cardinally and permutation variables, and b) a joint distribution over set elements with a fixed cardinality. Depending on the problem under consideration, we define different training models for set prediction using deep neural networks. % Since being unobservable from the annotations, we incorporate the permutation as latent discrete variable and estimate its distribution during the learning process using alternating optimization. 
We demonstrate the validity of our set formulations on relevant vision problems such as: 1) multi-label image
classification where we outperform the other competing methods on the PASCAL
VOC and MS COCO datasets, 2) object detection, for which our formulation outperforms popular state-of-the-art detectors, and 3) a complex CAPTCHA test, where we observe that, surprisingly, our set-based network acquired the ability of mimicking arithmetics without any rules being coded. 
\end{abstract}

}

\maketitle

% is not selected <OR> if conference mode is selected - because compsoc
% conference papers position the abstract like regular (non-compsoc)
\IEEEdisplaynontitleabstractindextext

% \fi
\IEEEpeerreviewmaketitle

\ifCLASSOPTIONcompsoc

\IEEEraisesectionheading{\section{Introduction}\label{sec:introduction}}
\else
\section{Introduction}
\label{sec:introduction}
\fi
\IEEEPARstart{D}{eep} structured networks such as deep convolutional (CNN) and recurrent (RNN) neural networks have enjoyed great success in many real-world problems, including scene classification~\cite{krizhevsky2012imagenet}, semantic segmentation~\cite{Papandreou:2015:ICCV}, speech recognition~\cite{hinton2012deep}, gaming~\cite{mnih2013playing,mnih2015human}, and image captioning~\cite{Johnson:2016:CVPR}. However, the current configuration of these networks is restricted to accept and predict structured inputs and outputs such as vectors, matrices, and tensors. %\footnote{Throughout the paper, we use the term \emph{structured data} for vectors, matrices and generally tensors which are fixed in size and ordered, \ie any permutation of its values changes its meaning. This is not to be confused with \emph{structured learning} which is commonly used for graphs.}. 
If the problem's inputs and/or outputs cannot be modelled in this way, all these learning approaches fail to learn a proper model~\cite{zaheer2017deep}. However, many real-world problems are naturally described as unstructured data such as sets~\cite{rezatofighi2018joint,zaheer2017deep}. A set is a collection of elements which is \emph{invariant under permutation} and the size of a set is \emph{not fixed} in advance.
%{Similar to sets, graphs in our definition can be assumed to be unstructured. Because the way that the nodes of a graph are ordered and represented can be changed while its topology remains the same. Therefore, they cannot be treated similar to structured data.}. 
Set learning using deep networks has generated substantial interest very recently~\cite{rezatofighi2017deepsetnet,rezatofighi2018joint,vinyals2015order,zaheer2017deep,murphy2018janossy,skianis2019rep, zhang2019deep}.
\begin{figure*}[htb!]
 	\centering
 	\includegraphics[width=.9\linewidth]{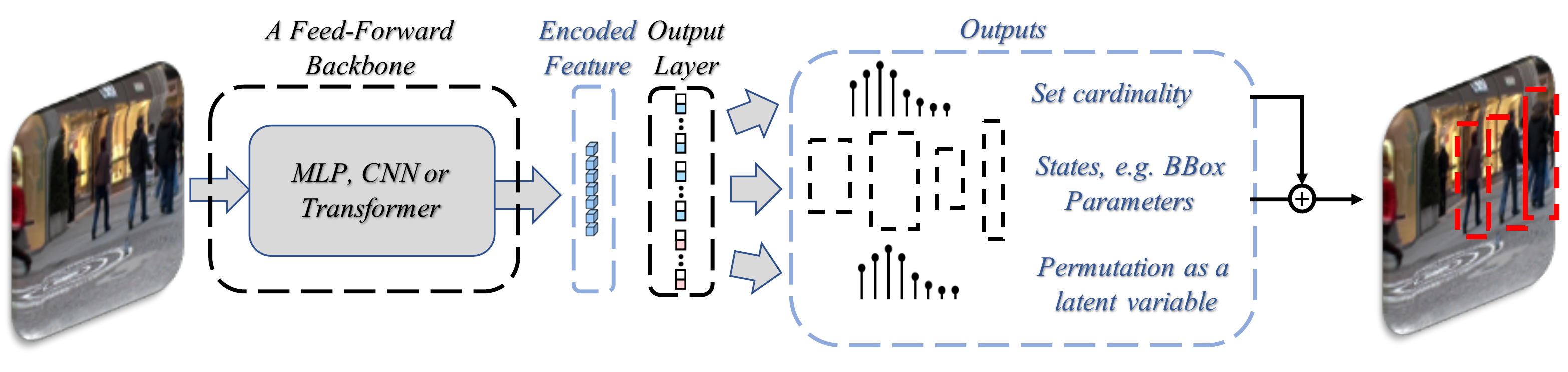}
 	\vspace{-0.50em}
 	% 'MOT16-RegProps', 'MOT16-ours', 'MOT16-MS-CNN'
 	\small
 	\caption{\small \revised{Prediction of a set, \eg an orderless set of bounding boxes with unknown cardinality, using a feed-forward neural network. A backbone, \eg a CNN, MLP or Transformer, first encodes an input image to a feature vector representation, which is then decoded by one/few output's layers (referred also as ``decoder'' in the text ) such as MLP or transformer decoder, into an output representation reflecting the states of the all set elements and their distributions over cardinality and permutation.}} 
 	\label{fig:teaserset}
 	\vspace{-1.0em}
 \end{figure*}

Consider the task of image classification as an example. The goal here is to predict a label (or a category) of a given image. The most successful approaches address this task with CNNs, \ie by applying a series of convolutional layers followed by a number of fully connected layers~\cite{krizhevsky2012imagenet,Simonyan:2014:VGG,Szegedy:2014:Inception,he2016deep}. The final output layer is a fixed-sized vector with the length corresponding to the number of categories in the dataset (\eg~1000 in the case of the ILSVR Challenge~\cite{Russakovsky:2015:ILSVRC}). Each element in this vector is a score or probability for one particular category such that the final prediction corresponds to an approximate probability distribution over all classes. The difficulty arises when the number of class labels is unknown in advance and in particular varies for each example. Then, the final output is generated heuristically by a discretization process such as choosing the $k$ highest scores~\cite{gong2013deep,wang2016cnn}, which is not part of the learning process. We argue that this problem should be naturally formulated as prediction of the sets of labels where the output does not have any known or fixed ordering or cardinality. As a second example, let us consider the task of object detection. Given a structured input, \eg an image as a tensor, the goal is to predict a set of \emph{orderless} locations, \eg bounding boxes, from an unknown and varying number of objects. Therefore, the output of this problem can be properly modelled as a set of entities. However, a typical deep learning network cannot be simply trained to directly predict a varying number of orderless locations. Existing \revised{popular} approaches formulate this problem using a pre-defined and fixed-sized grid~\cite{redmon2016you,redmon2017yolo9000,redmon2018yolov3,bochkovskiy2020yolov4} or anchor boxes~\cite{ren2015faster} representing a coarse estimate of all possible locations and scales of objects. Then, each location and scale is scored independently to determine whether or not it contains an object. The final output is generated heuristically by a discretization process such as non-maximum suppression (NMS), which is not part of the learning process. Therefore, their performance is hindered by this heuristic procedure. This is one reason why their solutions can only deal with moderate object occlusion. \revised{Few works~\cite{hosang2017learning,liu2019learning, hu2018relation} have also attempted to learn this process. However they assume the relationship between bounding boxes to be pairwise only. Moreover, an additional pairwise network and/or a classifier is introduced on top of the detection backbone to learn these pairwise relationships, introducing additional unnecessary computations to the problem.}
We argue that object detection should rather be posed as a set prediction problem, where a deep learning network is trained \revised{end-to-end} to output a \emph{set} of locations without any heuristics\footnote{\revised{Following our 2018 Arxiv work~\cite{rezatofighi2018deep}, two very recent frameworks~\cite{wang2021end, carion2020end} have adopted this new re-formulation, \ie end-to-end detection as a set prediction using \eg a set loss~\cite{carion2020end} and dynamic one-to-one assignment~\cite{wang2021end}, and have achieved state-of-the-art performance in the popular benchmark datasets.}}. 
%Another problem with this formulation is that the accuracy of localization is constrained by the resolution of this grid, For example, in most successful object detection approaches, the gird includes between $10K$ and $100K$ possible locations and scales~\cite{}, in order to detect just only few objects. This huge number of proposals make the training procedure inefficient and difficult~\cite{}. We argue that object detection problem can be formulated in principle as a set prediction problem. 

This shortcoming does not only concern object detection but also all problems where a set of instances as input and/or output is involved. Other examples of such tasks include predicting a set of topics or concepts in documents~\cite{blei2003latent}, sets of points in point cloud data~\cite{qi2017pointnet,qi2017pointnet++,  tchapmi2019topnet}, segmentation of object instances~\cite{lin2014microsoft} and a set of trajectories in multiple object tracking~\cite{babenko2011robust,alahi2016social,sadeghian2019sophie}. In contrast to problems such as classification, where the order of categories or the labels can be fixed during the training process and the output can be well represented by a fixed-sized vector, the instances are often unfixed in number and orderless. More precisely, it does not matter in which order the instances are labelled and these labels do not relate the instances to a specific visual category. To this end, training a deep network to input or predict instances seems to be non-trivial. We believe that set learning is a principled solution to all of these problems.%\revised{\footnote{\revised{Inspired by~\cite{rezatofighi2018deep,carion2020end} and with the new progresses toward stronger feed-forward neural models such as transformers~\cite{vaswani2017attention}, some of these problems are now reformulated as a set prediction task in the recent state-of-the-art frameworks~\cite{}}.}}. 

In this paper, we present a novel approach for learning to deal with sets using
deep learning. More clearly, in the presented model, we assume that the input (the observation) is still structured, \eg an image, but the annotated outputs are available as a set of labels \revised{(\Fig~\ref{fig:teaserset})}. This paper is a both technical and experimental extension of our recent works on set learning using deep neural networks~\cite{rezatofighi2017deepsetnet,rezatofighi2018joint}. 
Although our work in \cite{rezatofighi2017deepsetnet,rezatofighi2018joint} handles orderless outputs in the testing/inference step, it assumes that the annotated labels can be ordered in the learning step. This might be suitable for the problems such as multi-label image classification problem. However, many applications, such as object detection, the output labels cannot be naturally ordered in an structure way during training. Therefore, our model introduced in~\cite{rezatofighi2017deepsetnet,rezatofighi2018joint} cannot learn a sensible model for truly \revised{orderless} problems such as object detection  (see \Sec~\ref{result:od} for the experiment).
%The main limitation of that work, however, is that it does not \fixmeh{model the permutations of sets} in the learning step. Therefore, the method is only applicable to set problems that do not
%rely on permutation invariance during training, such as image
%tagging where the order of the labels can be fixed as a vector during training procedure. 
%However, when the order of the outputs cannot be fixed, \eg for the applications shown in this paper such as object detection, the proposed approach in~\cite{rezatofighi2018joint} cannot learn a sensible model to deal with the problem (see the supplementary material for the experiment). 
%\footnote{We performed an experiment for object detection using the set network proposed in~\cite{rezatofighi2018joint}. When attempting to train the network from orderless bounding boxes, the training objective does not converge (the details are provided in Appendix).}. 
In this paper in addition to what we proposed in ~\cite{rezatofighi2017deepsetnet,rezatofighi2018joint}, we also propose a complete set prediction formulation to address this limitation (\cf \textbf{Scenario 2} and \textbf{Scenario 3} in \Sec~\ref{sec:learning}). This provides a potential to tackle many other set prediction problems compared to~\cite{rezatofighi2017deepsetnet,rezatofighi2018joint}. To this end, in addition to multi-label image classification, we test this complete model on two new problems, \ie object detection and \revised{CAPTCHA} test.

Compared to our previous work~\cite{rezatofighi2017deepsetnet,rezatofighi2018joint}. the additional contribution of the paper is summarised as follows:\vspace{-0.7em}
\begin{enumerate}
	\setlength{\itemsep}{1pt}
	\setlength{\parskip}{0pt}
	\setlength{\parsep}{0pt}
	
	\item We provide a complete
	formulation for neural networks to deal with arbitrary sets with unknown permutation and cardinality as available annotated outputs. This makes a feed-forward neural network able to truly handle set outputs at both training and test time.
	
	\item Additionally, this new formulation allows us to learn the distribution over the unobservable permutation variables, which can be used to identify if there exist any natural ordering for the annotated labels. In some applications, there may exist one or several dominant orders, which are unknown from the annotations.
	 
	\item We reformulate object detection as a set prediction problem, where a deep network is learned end-to-end to generate the detection outputs with no heuristic involved. We outperform the popular state-of-the art object detectors on both simulated data and \revised{real data}.
	
	%\item Our approach can simultaneously detect and identify similar looking object instances across the test data due to the learning of the permutation variables, paving the path towards end-to-end training of a network for  multiple object tracking problem.
	 	
	\item We also demonstrate the applicability of our framework
	algorithm for a complex CAPTCHA test which can be formulated as a set prediction problem. 
\end{enumerate}

%\fixmel{This is indeed the YOLO way of doing things, but all the RCNN approaches use proposals, I think it is worth mentioning and adding the citations. I add the text now:}.
%Existing approaches formulate this problem by either using as a pre-defined and fixed-sized grid representing all possible locations and scales of the objects~\cite{redmond2016you}, or using object proposals which are then classified into the different classes~\cite{ren2015faster}.
%  \fixmeh{Faster r-cnn also has the same way, what is called Anchor boxes is actually a gird. They learned the abjectness score and then threshold it out and this is what you called proposal. all of the detectores are mapping to a grid. The difference between yolo and faster rcnn is that Yolo do object classification simoutanesouly with objectness score. Faster rcnn is doing in two steps }.

%This shortcoming concerns not only image tagging but also other problems like detection or graph optimization, where connectivity and graph size can be arbitrary. Similarly Dealing with graphs with arbitrary sizes and unfixed A arbitrary graph as inputs or outputs can be also seen as unstructured datain general.

%performing both learning and inference steps jointly. 

% !TEX root = main.tex

\section{Related Work}
\label{related work}
Handling unstructured input and output data, such as sets or point patterns, for both learning and inference is an emerging field of study that has generated substantial interest in recent years. Approaches such as mixture models~\cite{blei2003latent,hannah2011dirichlet,tran2016clustering}, learning distributions from a set of samples~\cite{muandet2012learning,oliva2013distribution}, model-based multiple instance learning~\cite{vo2017model}, and novelty detection from point pattern data~\cite{vo2016model},  can be counted as few out many examples that use point patterns or sets as input or output and directly or indirectly model the set distributions. However, existing approaches often rely on parametric models, \eg \iid Poisson point or Gaussian Process assumptions~\cite{adams2009tractable,vo2016model}.  Recently, deep learning has enabled us to use less parametric models to capture highly complex mapping distributions between structured inputs and outputs. This learning framework has demonstrated its great success in addressing pixel-to-pixel (or tensor-to-tensor) problems. There has been recently few attempts to apply this successful learning framework for the other types of data structures such as sets. One of earliest works in this direction
is the work of~\cite{vinyals2015order,Stewart:2016:CVPR}, which uses an RNN to read and predict sets. However, the output is still assumed to have a single order, which contradicts the orderless property of sets. Moreover, the framework can be used in combination with RNNs only and cannot be trivially extended to any arbitrary learning framework such as feed-forward architectures. Generally speaking, the
sequential techniques \eg RNN, LSTM and GRU, can deal with variable input or output sizes. However, they learn the parameters of a model for conditionally dependent inputs and outputs according to a chain rule; regardless if their orders assumed to be known or unknown during the training process~\cite{vinyals2015order}. To this end, these sequential learning techniques may not be an appropriate choice for encoding and decoding sets (verified also by our experiments). 

Alternatively the feed-forward neural networks has been recently attempted to be applied for inputting or outputting sets. 
Most existing approaches on set learning~\cite{zaheer2017deep,murphy2018janossy, skianis2019rep} have focused on the problem of encoding a set with a feed-forward neural network by using, \eg a shared network followed by a symmetrical pooling function~\cite{zaheer2017deep}, a permutation invariant pooling layer~\cite{murphy2018janossy} or a permutation invariant representation for sets~\cite{ skianis2019rep}. The similar concepts has been independently developed by the community working to encode point clouds using feed-forward neural networks~\cite{qi2017pointnet,qi2017pointnet++}. However, in all these works, the outputs of neural networks are either assumed to be a tensor or a set with the same entities of the input set, which prevents this approach to be used for the problems that require output sets with arbitrary entities. In this paper, we are instead interested in learning a model to output an arbitrary set. Somewhat surprisingly, there are only few works on learning to predict sets using deep neural networks~\cite{rezatofighi2017deepsetnet,rezatofighi2018joint, zhang2019deep}. The most related work to our problem is our previously proposed framework~\cite{rezatofighi2018joint} which seamlessly integrates a deep learning framework into set learning in order to learn to output sets. However, the approach only formulates the outputs with unknown cardinality and does not consider the permutation variables of sets in the learning step. 
Therefore, its application is limited to the problems with a fixed order output such as image tagging and diverges when trying to learn unordered output sets as for the object detection problem.
In this paper, we incorporate these permutations as unobservable variables in our formulation, and estimate their distribution during the learning process. 
Our unified set prediction framework has the potential to reformulate some of the existing problems, such as object detection, and to tackle a set of new applications, such as a logical CAPTCHA test which cannot be trivially solved by the existing architectures.

\section{Deep Perm-Set Network}
\label{JDSN}
%In this section, we first provide a mathematical definition of a set and tensor and demonstrate how a probability density function for a set can be modelled 
\begin{table*}[tb]\footnotesize
\centering
\caption{All mathematical symbols and notations used throughout the paper.}\vspace{-0.7em}
 \begin{tabular}{c c}
 \hline
\textbf{Symbol/Notation} & \textbf{Definition}
\\ 
\hline
\hline
$m$ & Cardinality variable \\
$\pi$ & Permutation variable\\
 $\bpi$ or $(\pi_1,\pi_2,\ldots,\pi_m)$ & Permutation vector for $m$ elements \\
 $\mathbb{N}^*$ & Space of all non-negative integer numbers\\
 $\bPi$ or $\{\bpi_1,\bpi_2,\cdots,\bpi_{m!}\}$  & Space  of  all  feasible  permutations for a vector with cardinality $m$\\
 \hline
$\calY$ or $\left\{\by_1,\cdots,\by_m\right\}$ & Set with \textbf{unknown} permutation and cardinality \\
$\calY^m$ or $\left\{\by_1,\cdots,\by_m\right\}^{m}$& Set with \textbf{unknown} permutation, but \textbf{known} cardinality \\
$\bY_{\bpi}^m$ or $\left(\by_{\pi_1},\cdots,\by_{\pi_m}\right)$ & Set with \textbf{known} permutation and cardinality (or a \textbf{tensor})\\
 $\bx$ & Input data as a tensor (\eg an RGB image)  \\
  $\bw$ & Collection of model parameters (\eg all deep neural network parameters)  \\
  $\D$ or $\{(\bx_{i},\calY_{i})\}$ & Training dataset\\
 \hline
 \hline
 $U$& Unit of  hyper-volume  in  the  feature  space\\
 $p(m)$ & Distribution over the cardinality variable $m$ (a discrete distribution)\\
  $p_m(\cdot)$ & Joint distribution of $m$ variables ($m$ is known)  \\
 \hline
 \end{tabular}
  \label{table:symbols}
\end{table*}

A set is a collection of elements which is invariant under permutation and the size of a set is not fixed in advance, \ie $\calY=\left\{\by_1,\cdots,\by_m\right\},  m\in\mathbb{N}^*$.
 A statistical function describing a finite-set variable $\calY$ is a
 combinatorial probability density function $p(\calY)$ defined by $p(\calY) = p(m)U^m p_m(\{\by_{1},\by_{2},\cdots,\by_{m}\}),$
where $p(m)$ is the cardinality distribution of the set $\calY$ and  $p_m(\{\by_{1},\by_{2},\cdots,\by_{m}\})$ is a symmetric joint probability density distribution of the set given known cardinality $m$. $U$ is the unit of hyper-volume in the feature space, which
 cancels out the unit of the probability density
 $p_m(\cdot)$ making it unit-less, and thereby avoids the unit
 mismatch across the different dimensions (cardinalities)~\cite{vo2017model}.

Throughout the paper, we use $\calY=\left\{\by_1,\cdots,\by_m\right\}$ for a set with  unknown cardinality and permutation, $\calY^m=\left\{\by_1,\cdots,\by_m\right\}^{m}$ for a set with known cardinality $m$ but unknown permutation and $\bY_{\bpi}^m=\left(\by_{\pi_1},\cdots,\by_{\pi_m}\right)$ for an ordered set with known cardinality (or dimension) $m$ and permutation $\bpi$, which means that the $m$ set elements are ordered under the permutation vector $\bpi = (\pi_1,\pi_2,\ldots,\pi_m)$. Note that an ordered set with known dimension and permutation exactly corresponds to a tensor, \eg  a vector or a matrix.

According to the permutation invariant property of the sets, the set $\calY^m$ with known cardinality $m$ can be expressed by an ordered set with any arbitrary permutation, \ie $\calY^m := \{\bY^m_{\bpi}| \forall \bpi\in \bPi\}$, where, $\bPi$ is the space of all feasible permutation $\bPi=\{\bpi_1,\bpi_2,\cdots,\bpi_{m!}\}$ and $|\bPi|= m!$. Therefore, the probability density of a set $\calY$ with unknown permutation and cardinality conditioned on the input $\bx$ and the model parameters $\bw$ is defined as \vspace{-.2em}
\begin{equation}
\begin{aligned}
p(\calY|\bx,\bw)\!\! & = \!p(m|\bx,\bw)\!\times\! U^m\!\! \times\! p_m(\calY^m|\bx,\bw),\\
\!\!& = \!p(m|\bx,\bw)\!\times \!U^m\!\! \times\!\!\!\!\!\sum_{\forall\bpi\in \bPi}\!\!\!p_m(\bY^m_{\bpi},\bpi|\bx,\bw).
%&= p(m|\bx,\bw)\times U^m \times\sum_{\forall\bpi\in \bPi}\bigg[p_m(\bpi|\bx,\bw)\times p(\bY^m_{\bpi}|\bx,\bw,\bpi)\bigg].
\label{eq:setdist}
\vspace{-.2em}
\end{aligned}
\end{equation} 
The parameter vector $\bw$ models both the \emph{cardinality} distribution of the set elements $p(m|\cdot)$ as well as the joint state distribution of set elements and their \emph{permutation} for a fixed cardinality $p_m(\bY^m_{\bpi},\bpi|\cdot)$.

The above formulation represents the probability density
of a set which is very general and completely independent
of the choices of cardinality, state and permutation distributions. 
It is thus straightforward to transfer it to many applications
that require the output to be a set. Definition of these distributions for the applications in this paper will be elaborated later.

%However, to
%make the problem amenable to mathematical derivation and
%implementation, we adopt the following assumptions: \emph{i)} the outputs in the set are derived from an independent
%and identically distributed (\iid)-cluster point process model, and \emph{ii)} their cardinality follows
%a categorical distribution parameterised by event probabilities $\brho$.
% \fixmea{ok, I still find it a bit strange that you say parameter $\brho$ and two lines later $\brho$ is the vector of probabilities}\fixmeh{The parameters for a categorical distribution called the event probabilities and $\brho$ is the parameter for the categorical distribution. still I don't know what is confusing here for you}
% ok, didnt know the parameter was called scent probabilities
%Thus, we can write the distribution
%as
%\begin{equation}
%\begin{aligned}
%p(\calY|\bw,\bx) = \int p(m|\brho)&p(\brho|\bx,\bw) d\brho \times  U^m \times\sum_{\bpi\in \bPi}\bigg(p_m(\bpi|\bx,\bw) p_m(\bY^m_{\bpi}|\bx,\bw,\bpi)\bigg)\\ 
%% = \int p(m|\brho)&p(\brho|\bx,\bw) d\brho \times  U^m \times\sum_{\bpi\in \bPi}\left(p_m(\bpi|\bx,\bw)\times\prod_{\substack{\by_{\pi}\in\bY^m_{\bpi}}} p(\by_{\pi}|\bx,\bw)\right),
%\label{eq:posterior_general} 
%\end{aligned}
%\end{equation} 
%where $p(\by_{\pi}|\cdot,\cdot)$ denotes the probability of taking on the state $\by$ in a specific order $\pi$, and $\brho = (\rho_1,\ldots,\rho_M)$ is the vector of event probabilities, 
%\emph{i.e.} $\sum_{i=1}^M \rho_i = 1$ and $\rho_i>0,\forall i\in\{1,\ldots,M\}$.

\subsection{Posterior distribution}
\label{sec:posterior}

Let $\D = \{(\bx_{i},\calY_{i})\}$ be a training set,
where each training sample $i=1,\ldots,n$ is a pair consisting of an input feature (\eg image), $\bx_{i}\in\mathbb{R}^{l}$ and an output set
$\calY_{i} = \{\by_{1},\by_{2},\ldots,\by_{m_i}\}, \by_{k}\in\mathbb{R}^{d}, m_i\in\mathbb{N}^*$.
The aim is now to learn the parameters $\bw$ to estimate the set distribution in Eq. (\ref{eq:setdist}) using the training samples. 

To learn the parameters $\bw$, we assume that the training samples are independent from each other and the distribution $p(\bx)$ from which the input data is sampled is independent from both the output and the parameters. 
Then, the posterior distribution over the parameters can be derived as\vspace{-.2em} %$\displaystyle p(\bw|\D)\propto p(\D|\bw)p(\bw)\\\propto\prod_{i=1}^{n}\Bigg[ p(m_{i}|\bx_{i},\bw)\times 
%U^{m_i}\times\sum_{\forall\bpi\in\bPi}p_m(\bpi|\bx_i,\bw)\\ \times p_m(\bY_{\bpi}^{m_i}|\bx_i,\bw,\bpi)\Bigg]p(\bw).$
\begin{equation*}
\begin{aligned}
p(\bw|\D)&\propto p(\D|\bw)p(\bw)\\&\propto\prod_{i=1}^{n}\Bigg[ p(m_{i}|\bx_{i},\bw)\times 
U^{m_i}\times\sum_{\forall\bpi\in\bPi}p_m(\bpi|\bx_i,\bw)\\&\qquad\qquad\qquad\qquad\times p_m(\bY_{\bpi}^{m_i}|\bx_i,\bw,\bpi)\Bigg]p(\bw).
\vspace{-1.5em}
\end{aligned}
\label{eq:posterior}
\end{equation*}
Note that $p_m(\bY^{m_i}_{\bpi},\bpi|\cdot)$ is decomposed according to the chain rule and $p(\bx)$ is eliminated as it appears in both the numerator and the denominator. We also assume that the outputs in the set are derived from an independent and identically distributed (\iid) cluster point process model.
%To make the problem amenable to mathematical derivation and implementation, we follow the same assumptions used in~\cite{rezatofighi2018joint}, \ie \emph{i)} the outputs in the set are derived from an independent and identically distributed (\iid)-cluster point process model, and \emph{ii)} their cardinality follows a categorical distribution parameterised by event probabilities $\brho$, \ie
%\begin{eqnarray*}
%	m & \sim & \text{Cat}(m;\brho)\\
%	\brho & \sim & \text{Dir}(\brho;\balpha(\bx,\bw)).
%\end{eqnarray*}
%Here $\balpha(\bx,\bw) = [\alpha_i (\bx,\bw)]_{i=1}^m $ are the parameters of the Dirichlet distribution above, and $\alpha_i (\bx,\bw) > 0 $ for all $\bx,\bw$ and all $i = 1, ..., m$.    
Therefore, the full posterior distribution can be written as \vspace{-.5em}
\begin{equation}
\begin{aligned}
&p(\bw|\D) \propto\prod_{i=1}^{n}\Bigg[p(m_{i}|\bx_{i},\bw)\times 
U^{m_i}\times \\&\sum_{\forall\bpi\in\bPi}\!\!\!\left(p_m(\bpi|\bx_{i},\bw)\times\!\!\!\prod_{\sigma=\pi_1}^{\pi_{m_i}}p(\by_{\sigma}|\bx_{i},\bw,\bpi)\right) \Bigg]p(\bw).
\label{eq:full-posterior}
\end{aligned}   
\end{equation}

%where $DC$ is a Dirichlet-Categorical distribution defined by 
%\begin{equation}
%DC\left(m;\balpha\right)  =  \frac{\alpha_m+C_m}{\sum_{\acute{m}}\alpha_{\acute{m}}+C}\\, 
%\label{eq:DC}
%\end{equation}
%where $C_m$ is the number of samples in the training set with cardinality $m$, and $C$ is the total number of training samples.

%In this paper, we use two categorical distributions to define cardinality $p(m_{i}|\cdot,\cdot)$ and permutation $p_m(\bpi|\cdot,\cdot)$ terms. However depending of the application, any discrete distribution such as Poisson, binomial, negative binomial or Dirichlet-categorical (\cf~\cite{rezatofighi2017deepsetnet,rezatofighi2018joint}, can be used for these terms. Moreover, we find the assumption about \iid cluster point process practical for the reported applications. Nevertheless, the extension to non-\iid cluster point process model for any other application would be a potential research direction for this work. 

We would like to re-emphasize that the assumption here is that the annotated labels, $\calY_i$, are available as a set of entities, \eg a set of image tags or bounding boxes, and this means that we might not have any knowledge if there exist any natural ordering structure in the data. Our proposed solution is capable of inferring these potential ordering structures (if they exist at all) in the data by learning the distribution over the permutations, \ie $p_m(\bpi|\cdot,\cdot)$. We can also assume that $p_m(\bpi|\cdot,\cdot)$ is uniform over all permutations (\ie the order does not matter) and learn the model accordingly. Moreover, in some applications, \eg image tagging, we can assume that the permutation of the outputs for all the training data for the training stage can be consistently fixed. Therefore, in this case the permutation is not a random variable and will be eliminated from the posterior distribution. The learning process for each of these scenarios is slightly different, therefore, in the next section we detail them as Scenarios 1, 2 and 3.

%According to the ranking function defined in~\cite{}, we can replace the above distribution by a proper unitless distribution function independent of the unit of hyper-volume $U$ as
%\begin{equation}
%\begin{aligned}
%p(\bw|\D)  \propto\prod_{i=1}^{n}\bigg[DC\left(m_{i};\balpha(\bx_{i},\bw)\right)\times \left(\prod_{y\in\calY_i}\frac{p(y|\bx_i,\bw)}{\parallel p(y|\bx_i,\bw)\parallel_2^2}\right)\bigg]p(\bw),
%\label{eq:full-posterior-final}
%\end{aligned}   
%\end{equation}
%where $\parallel p(y|\cdot)\parallel_2^2
%=
%\int
%p^2(y|\cdot)
%dy$ is the squared L2-norm of $p(y|\cdot)$.

\subsection{Learning}
\label{sec:learning}

For learning the parameters in the aforementioned scenarios, we use a point estimate for the posterior,
%\footnote{The extension to its Bayesian estimation can be explored as a potential future work.},
\textit{i.e.} $p(\bw|\D) = \delta(\bw=\bw^{*}|\D)$, where $\bw^{*}$ is computed using the MAP estimator, \ie $\bw^{*} = \arg\min_{\bw}\enspace -\log\left(p\left(\bw|\D\right)\right)$. Since $\bw$ in this paper is assumed to be the parameters of a feed-forward deep neural network, to estimate $\bw^{*}$, we use commonly used stochastic gradient decent (SGD), where one iteration is performed as $\bw_{k} = \bw_{k-1}-\eta \frac{-\partial \log\left(p\left(\bw_{k-1}|\D\right)\right)}{\partial\bw_{k-1}},$ where $\eta$ is the learning rate. 
\subsubsection{\bf Scenario 1: Permutation can be fixed during training}
\label{sec:scenario1}
In some problems, although the output labels  represent a set (without any preferred ordering), during training step they can be consistently ordered in an structure way. For example, in multi-label  image  classification problem, all possible tags can be ordered in an arbitrary way, \eg (Person, Dog, Bike,$\cdots$) or (Dog, Person, Bike, $\cdots$), but they can be fixed to be the exactly same order for all training instances during training step. 

In this scenario, since $\pi$ is not a random variable in this case,  $p(\by_{\sigma}|\bx_{i},\bw,\bpi)= p(\by_{\sigma}|\bx_{i},\bw)$. Therefore the term $p_m(\bpi|\cdot,\cdot)$ is marginalized out from the posterior distribution
%\vspace{-1.1em}
\begin{equation}
\begin{aligned}
p(\bw|\D) &\propto\prod_{i=1}^{n}\Bigg[p(m_{i}|\bx_{i},\bw)\times 
U^{m_i}\times \\&\prod_{\sigma=1}^{{m_i}}p(\by_{\sigma}|\bx_{i},\bw)\!\times\!\!\underbrace{\sum_{\forall\bpi\in\bPi}\!\!\!\left(p_m(\bpi|\bx_{i},\bw)\right)}_{=1} \Bigg]p(\bw),
\label{eq:full-posterior-s1}
\end{aligned} 
\end{equation}
 and it is simplified as 
%\vspace{-1.5em}
\begin{equation}
\begin{aligned}
p(\bw|\D)\propto\prod_{i=1}^{n}\Bigg[p(m_{i}|\bx_{i},\bw)&\times
U^{m_i}\times
\\&\prod_{\sigma=1}^{{m_i}}p(\by_{\sigma}|\bx_{i},\bw) \Bigg]p(\bw).
\label{eq:full-posterior-s1b}
\end{aligned}   
\end{equation}
%For simplicity, we use a point estimate for the posterior $p(\bw|\D)$, \textit{i.e.} $p(\bw|\D) = \delta(\bw=\bw^{*}|\D)$, where $\bw^{*}$ is computed using the MAP estimator, \ie $\bw^{*} = \arg\max_{\bw}\enspace \log\left(p\left(\bw|\D\right)\right)$.
Therefore, for learning the parameters of a neural network $\bw$, we have  
\begin{equation}
\begin{aligned}
\bw^{*}  = &\arg\min_{\bw}\enspace -\log\left(p\left(\bw|\D\right)\right),\\
 =  
&\arg\min_{\bw}\enskip\sum_{i=1}^{n}\bigg[\underbrace{-\log\left(p(m_{i}|\bx_{i},\bw)\right)}_{\calL_{card}(\cdot)}-\underbrace{\cancel{m_{i}\log U}}_{\text{removed}}\\
&\qquad+\sum_{\sigma=1}^{{m_i}}\bigg(\underbrace{-\log\big(p(\by_{\sigma}|\bx_{i},\bw)\big)}_{\calL_{state}(\cdot)}\bigg)\bigg]+\gamma\|\bw\|_2^2,
\label{eq:map_complete}
\end{aligned}
\end{equation}
\begin{equation}
\begin{aligned}
\bw^{*}   =  
&\arg\min_{\bw}\enskip \sum_{i=1}^{n}\bigg[\calL_{card}\left(m_{i}, \balpha\left( \bx_{i},\bw\right)\right)+\\&\qquad\sum_{\sigma=1}^{{m_i}}\bigg(\calL_{state}(\by_{\sigma},\bO^{\sigma}_1\left( \bx_{i},\bw\right))\bigg)\bigg]+\gamma\|\bw\|_2^2.
\label{eq:map_complete_S1}
\end{aligned}
\end{equation}
where $\gamma$ is the regularisation parameter, which is also known as the weight decay parameter and is commonly used in training neural networks. $\calL_{card}(\cdot,\balpha (\cdot,\cdot))$ and $\calL_{state}(\cdot,\bO^{\sigma}_1 (\cdot,\cdot))$ represent cardinality and state losses, respectively, where $\balpha (\cdot,\cdot)$ and $\bO^{\sigma}_1 (\cdot,\cdot)$ are respectively the part of output layer of the  neural network, which predict the cardinality and the state of each of $m_i$ set elements. 

\textbf{Implementation insight:} In this scenario, the training procedure is simplified to jointly optimize  cardinality $\calL_{card}(\cdot,\balpha (\cdot,\cdot))$ and state $\calL_{state}(\cdot,\bO^{\sigma}_1 (\cdot,\cdot))$ losses under a pre-fixed permutation of output labels~\footnote{This is only possible for some applications, where the outputs are category or class labels rather than instance labels, \eg multi-label image classification.}. 

The \revised{cardinality} loss, $\calL_{card}(\cdot,\balpha (\cdot,\cdot))$, can be the negative log of any discrete distribution such as Poisson, binomial, negative binomial, categorical (softmax) or Dirichlet-categorical (\cf~\cite{rezatofighi2017deepsetnet,rezatofighi2018joint,rezatofighi2018deep}). Therefore, $\balpha (\cdot,\cdot)$ represents the parameters of this discrete distribution, \eg a single parameter for Poisson or $M+1$ parameters (\ie $M+1$ event probabilities for set with maximum cardinality $M$) for a categorical distribution. We will compare some of those cardinality losses for the image tagging problem in \Sec~\ref{sec:results}.

For the state loss, $\calL_{state}(\cdot,\bO^{\sigma}_1 (\cdot,\cdot))$, with variable output set cardinality, we assume that the maximum set cardinality to be generated is known. This would be a practical assumption for some applications, \eg image tagging, where the output set is always a subset of the maximum sized set with all pre-defined $M$ labels, \ie  $\calY\subseteq\{\ell_1,\ell_2,\cdots,\ell_M\}$. Therefore, the output of the network for this part would be the collection of the states and existence score for all $M$ set elements, \ie $\bO_1 = (\bO^1_1,\bO^2_1,\cdots,\bO^M_1)$. In image tagging, the state loss, $\calL_{state}(\cdot,\bO^{\sigma}_1 (\cdot,\cdot))$ can be simply a loss defined over the predicted existence score for each $M$ labels in an image, \eg a logistic regression or binary cross entropy (BCE) loss. Remember that, in this scenario, the assignment of the truth (labels in image tagging) to the index of the outputs is known and fixed during training.

% we can rewrite an equivalent  binary formulation for the above MAP problem as
% \begin{equation}
% \begin{aligned}
% \bw^{*}=   &\arg\max_{\bw}\enskip\sum_{i=1}^{n}\bigg[\sum_{\ell=1}^M z_i^{\ell}\log p(z_i^{\ell}|\bx_i,\bw)+\sum_{\ell=1}^M z^{\ell}_i \log U\\
% & +\log DC\left(\sum_{\ell=1}^M z^{\ell}_i ;\balpha(\bx_{i},\bw)\right)\bigg]-\gamma\|\bw\|_2^2\\
% =&\arg\max_{\bw}\enskip\sum_{i=1}^{n}\bigg[\sum_{\ell=1}^M z_i^{\ell}\log p(z_i^{\ell}|\bx_i,\bw)+
% \\&\log DC\left(\sum_{\ell=1}^M z^{\ell}_i ;\balpha(\bx_{i},\bw)\right)\bigg]-\gamma\|\bw\|_2^2,
% \end{aligned}
% \label{eq:map_dual}
% \end{equation}
% where $z_i^{\ell}\in\{0,1\}$ represents the existence or non-existence of any specific label in the image $\bx_i$.  $p(z_i^{\ell}=1|\bx_i,\bw)$ can be defined as a binary logistic regression function  
% \begin{equation*}
% p(z_i^{\ell}=1|\bx_i,\bw) = \frac{\exp{O^\ell(x_i,\bw)}}{1+\exp{O^\ell(x_i,\bw)}},
% \label{eq:BCE_loss}
% \end{equation*}
% where $O^{\ell}(x_i,\bw)$ is the network's predicted output corresponding to the $\ell^\text{th}$ label.\\\\
\subsubsection{\bf Scenario 2: Learning the distribution over the permutations} In the problems, where the outputs are set of instances, instead of a set of categories or classes, similar to the previous scenario, we cannot create a consistent fixed ordering representation for all instances during training procedure.

In this scenario, we assume that the permutation cannot be fixed at training time. Therefore, the posterior distribution is exactly as defined in \Eq~(\ref{eq:full-posterior}). For learning the parameters and in order to optimize the negative logarithm of the posterior over the parameters, \ie $-\log\left(p\left(\bw|\D\right)\right)$, we need to marginalize over all permutations in every iteration of SGD, which is combinatorial and can become intractable even for relatively small-sized problems. To address this, we approximate this marginalization with the most significant permutations (the permutation samples with high probability) for each training instance from the samples attained in each iteration of SGD, \ie \vspace{-0.5em} 
\revised{\begin{equation}
\begin{aligned}
p_m(\bpi|\bx_{i},\bw) = &\sum_{\forall \hat{\bpi}_{i}\in\bPi} \omega_{\hat{\bpi}_{i}}(\bx_{i},\bw)\delta[\bpi-\hat{\bpi}_{i}]\\ \approx& \frac{1}{N_\kappa}\sum_{\forall\bpi^{*}_{i,k}\in\bPi} \tilde{\omega}_{\bpi^{*}_{i,k}}(\bx_{i},\bw)\delta[\bpi-\bpi^{*}_{i,k}],
\label{eq:permutation dist}
\end{aligned}
\end{equation}}
\revised{where $\delta[\cdot]$ is the Kronecker delta and $\sum_{\forall \hat{\bpi}\in\bPi} \omega_{\hat{\bpi}}(\cdot,\cdot) = 1$. $\hat{\bpi}_{i}$ is one of the all possible permutations for the training instance $i$, and $\bpi^{*}_{i,k}$ is the most significant permutation for the training instance $i$}, sampled from $p_m(\bpi|\cdot,\cdot)\times\prod_{\sigma=\pi_1}^{\pi_{m_i}}p(\by_{\sigma}|\cdot,\cdot,\bpi)$ during $k^{th}$ iteration of SGD (using Eq.~\ref{eq:alternation1}). The weight $\tilde{\omega}_{\bpi^{*}_{i,k}}(\cdot,\cdot)$ is proportional to the number of the same permutation samples $\bpi^{*}_{i,k}(\cdot,\cdot)$, extracted during all SGD iterations for the training instance $i$ and $N_\kappa$ is the total number of SGD iterations. Therefore, $\sum_{\forall\bpi^{*}_{i,k}\in\bPi} \tilde{\omega}_{\bpi^{*}_{i,k}}(\cdot,\cdot)/ N_\kappa = 1$.  Note that at every iteration, as the parameter vector $\bw$ is updated, the best permutation $\bpi^{*}_{i,k}$ can change accordingly even for the same instance $\bx_i$. This allows the network to traverse through the entire space $\bPi$ and to approximate $p_m(\bpi|\bx_{i},\bw)$ by a set of significant permutations. To this end, $p_m(\bpi|\bx_{i},\bw)$ is assumed to be point estimates for each iteration of SGD. Therefore, \vspace{-0.5em} 
\begin{equation}
\begin{aligned}
p(\bw_k|\D) \!\!\!\!\!\!\!\appropto\!\!\!\!\!\!\prod_{i=1}^{n}\!&\Bigg[p(m_{i}|\bx_{i},\bw_k)\!\times 
\!U^{m_i}\!\!\times\!\tilde{\omega}_{\bpi^{*}_{i,k}}\!(\bx_{i},\bw_k) \\&\quad\times\!\!\!\prod_{\sigma=\pi_1}^{\pi_{m_i}}p(\by_{\sigma}|\bx_{i},\bw_k,\bpi^*_{i,k}) \Bigg]p(\bw_k).
\label{eq:appr-posterior}
\end{aligned}   
\end{equation}

In summary, to learn the parameters of the network $\bw$, the best permutation sample for each instance $i$ in each SGD iteration $k$, \ie $\bpi_{i,k}^*$ is attained by solving the following assignment problem, first: 
\begin{equation}
\label{eq:alternation1}
\begin{aligned}
\bpi_{i,k}^*&=\arg\min_{\bpi\in\bPi}  -\log\!\bigg(\!p_m(\bpi|\cdot,\cdot)\!\times\!\!\!\!\prod_{\sigma=\pi_1}^{\pi_{m_i}}\!\!p(\by_{\sigma}|\cdot,\cdot,\bpi)\!\bigg)\\&=\arg\min_{\bpi\in\bPi} \quad \calL_{perm}\bigg(\bpi,\bO_2(\bx_{i},\bw_{k-1})\bigg) \\&\qquad\qquad\quad +\sum_{\sigma=\pi_1}^{\pi_{m_i}}\bigg(\calL_{state}\big(\by_{\sigma},\bO^{\sigma}_1(\bx_{i},\bw_{k-1})\big)\bigg) ,
\end{aligned}
\vspace{-0.5em}
\end{equation}
and then use the sampled permutation $\bpi_{i,k}^*$ to apply standard back-propagation as follows:
% in each iteration of SGD using  in order to compute the best permutation sample in each iteration of SGD, $\bpi^*_{i,k}$, we first find the best permutation sample solve an assignment problem using

% To compute $\bw_k$ and $\bpi^*_{i,k}$, we use alternating optimization and use standard backpropagation
% \begin{equation}
% \label{eq:alternation1}
% \begin{aligned}
% \bpi_{i,k}^*=\arg\min_{\bpi\in\bPi} \quad &f_1\bigg(\bY^{m_i}_{\bpi},\bO_1(\bx_{i},\bw_{k-1})\bigg)\\&+f_2\bigg(\bpi,\bO_2(\bx_{i},\bw_{k-1})\bigg),
% \end{aligned}
% \vspace{-0.5em}
% \end{equation}

\begin{figure*}[t]
 	\centering
 	\includegraphics[width=.9\linewidth]{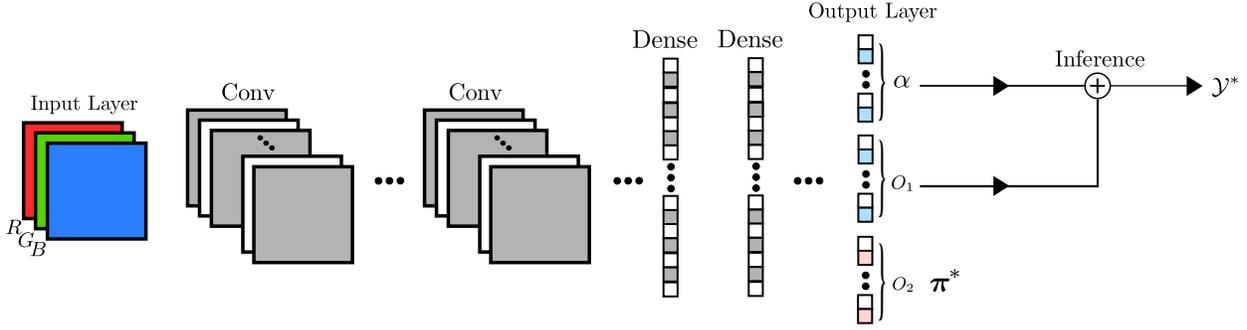}
 	\vspace{-0.50em}
 	% 'MOT16-RegProps', 'MOT16-ours', 'MOT16-MS-CNN'
 	\small
 	\caption{\small A schematic overview of our Deep Perm-Set Network (\textbf{Scenario 2}). \revised{A tensor input, \eg an RGB image, is passed through a backbone, \eg  CNN, and output layers, \eg series of fully connected layers with a collection of all parameters denoted as $\bw$. The backbone can be either one of the standard convolutional backbones~\cite{krizhevsky2012imagenet, Simonyan:2014:VGG, he2016deep, huang2017densely}, a transformers encoder~\cite{vaswani2017attention} or both and the output layers can be a stack of fully connected layers or a transformers decoder~\cite{vaswani2017attention}.} The final output layer consists of three parts shown by $\balpha$, $\bO_1$ and $\bO_2$, which respectively predict the cardinality, the states and the permutation of the set elements. During training, $\bpi^*_{i,k}$ representing a permutation sample (attained by Eq.~\ref{eq:alternation1}), is used as the ground truth to update the loss $\calL_{perm}(\bpi^*_{i,k},\bO_2)$ and to sort the elements of the ground truth set's state in the $\calL_{state}(\bY^{m_i}_{\bpi^*_{i,k}},\bO_1)$ term in Eq.~\ref{eq:alternation2}. During inference, the optimal set $\calY^{*}$ is only calculated using the cardinality $\balpha$ and the states $\bO_1$ outputs. $\bpi^{*}$ is an extra output for ordering representation. For \textbf{Scenarios 1 and 3}, the permutation head, \ie $\bO_2$, is not defined.} 
 	\label{fig:setcnn}
 	\vspace{-1.0em}
 \end{figure*}

% use alternating optimization and use standard backpropagation
% to learn the parameters of the deep neural network. \vspace{-0.5em}  
% \begin{equation}
% \label{eq:alternation1}
% \begin{aligned}
% \bpi_{i,k}^*=\arg\min_{\bpi\in\bPi} \quad &f_1\bigg(\bY^{m_i}_{\bpi},\bO_1(\bx_{i},\bw_{k-1})\bigg)\\&+f_2\bigg(\bpi,\bO_2(\bx_{i},\bw_{k-1})\bigg),
% \end{aligned}
% \vspace{-0.5em}
% \end{equation}
\begin{equation}
\label{eq:alternation2}
\begin{aligned}
\bw_{k}& = \bw_{k-1}-\eta\frac{-\partial \log\left(p\left(\bw_{k-1}|\D\right)\right)}{\partial\bw_{k-1}}\\
& = \bw_{k-1}-\eta\sum_{i=1}^{n}\Big[\frac{\partial \calL_{card}\big(m_i,\balpha\big)}{\partial\balpha}.\frac{\partial \balpha}{\partial\bw}+\\&\qquad\qquad\qquad\frac{\partial \calL_{perm}\big(\bpi_{i,k}^*,\bO_2\big)}{\partial\bO_2}.\frac{\partial \bO_2}{\partial\bw}+\\&\qquad\qquad\sum_{\sigma=\pi^*_1}^{\pi^*_{m_i}}\frac{\partial \calL_{state}\big(\by_{\sigma},\bO^{\sigma}_1\big)}{\partial\bO^{\sigma}_1}.\frac{\partial \bO^{\sigma}_1}{\partial\bw}\!\Big]+ 2\gamma\bw,
\end{aligned}
\end{equation}
where $\calL_{card}(\cdot,\balpha (\cdot,\cdot))$, $\calL_{perm}(\cdot,\bO_2 (\cdot,\cdot))$ and $\calL_{state}(\cdot,\bO^{\sigma}_1 (\cdot,\cdot))$ respectively represent cardinality permutation and state losses, which are defined on the part of output layer of the neural network respectively representing the cardinality, $\balpha (\cdot,\cdot)$, the permutation, $\bO_2 (\cdot,\cdot)$, and the state of each of $m_i$ set elements, $\bO^{\sigma}_1 (\cdot,\cdot)$ (\cf~Fig.~\ref{fig:setcnn}). 
% where  $\gamma$ is the regularization parameter, $f_1\big(\bY^{m_i}_{\bpi},\bO_1(\bx_{i},\bw)\big) =-\sum_{\sigma=\pi_1}^{\pi_{m_i}} \log\big(p(\by_{\sigma}|\bx_{i},\bw,\bpi)\big)$,
% \\$f_2\big(\bpi,\bO_2(\bx_{i},\bw)\big) =-\log\big(\tilde{\omega}_{\bpi}\big(\bx_{i},\bw\big)\big)$, and $f_3\big(m_i,\balpha(\bx_{i},\bw)\big) = -\log\left(p(m_{i}|\bx_{i},\bw)\right)$,  
% defined on the part of output layer of the neural network representing the cardinality, $\balpha (\cdot,\cdot)$, the permutation, $\bO_2 (\cdot,\cdot)$, and the state of each of $m_i$ set elements, $\bO^{\sigma}_1 (\cdot,\cdot)$ (Fig.~\ref{fig:setcnn}).

\textbf{Implementation insight:} In this scenario, for training neural network parameters $\bw$, we need to jointly optimize three losses: cardinality $\calL_{card}(\cdot,\balpha (\cdot,\cdot))$, permutation $\calL_{perm}(\cdot,\bO_2 (\cdot,\cdot))$ and state $\calL_{state}(\cdot,\bO^{\sigma}_1 (\cdot,\cdot))$, using SGD while the assignment (permutation) between ground truth annotations and the network outputs labels are unknown and interchangeable, which is sampled in each iteration by solving Eq.~\ref{eq:alternation1}. Note that this equation is a discrete optimization to find the best permutation $\bpi^*_{i,k}$, which can be solved using any independent discrete optimization approach depending on the choice of the objectives. In our paper, we use Hungarian (Munkres) algorithm. 

Similar to \textbf{Scenario 1},  $\balpha(\cdot,\cdot)$ represents the parameters of the cardinality distribution (a discrete distribution) \ie $\bO_1 = (\bO^1_1,\bO^2_1,\cdots,\bO^M_1)$ are the state parts of the network output representing the collection of the states and existence score for all $M$ set elements with the maximum size. For example in object detection, each $\bO^1_1$ can simply represent parameters and existence for an axis-aligned bounding box, \eg $(x,y,w,h,s)$. However, in contrast to \textbf{Scenario 1}, the assignment (permutation) between ground truth annotations and the network outputs cannot be pre-fixed during training and \revised{it is dynamically estimated from the best permutation samples, \ie $\bpi^*_{i,k}$}, in each SGD iterations and for each input instance. Compared to \textbf{Scenario 1}, which is a proper set learning model for predicting a set of tags or category labels, this (and also the next scenario) is a proper model for learning to output a set of instances, \eg bounding boxes, masks and trajectories, where the assignment between ground truth annotations and the network outputs can not be fixed; otherwise the network does not learn a sensible model for the task (See the supp. material document for the experiment). 

In addition, in this scenario the outputs $\bO_2 (\cdot,\cdot)$ represent the parameters of another discrete distribution (the permutation distribution), where each discrete value in this distribution defines a unique permutation $\bpi$ of $M$ set elements (the biggest set) and its ground truth for each input instance in each SGD iteration is attained from Eq.~\ref{eq:alternation1}, \ie $\bpi^*_{i,k}$. Note that the ground truth annotations are assumed to be a set, \ie we do not have any preference about their ordering or we do no have any prior knowledge if there is indeed any specific ordering structure in the annotations. By learning from the sampled permutations during SGD, we can have an extra information about the ordering structure of the annotations, \ie there exist a single or multiple orders which matter or the problem is truly orderless. 
% \fixmea{This last paragraph is quite confusing. What do the last two sentences say?}

% Finally, $\bpi^*_{i,k}$ representing a permutation sample, is used as the ground truth to update $f_2(\cdot)$ and to sort the elements of the ground truth set's state in the $f_1(\cdot)$ term in Eq.~\ref{eq:alternation2}. By learning $f_2(\cdot)$, we indeed learn the model for the output $\bO_2 (\cdot,\cdot)$ representing the weights of the permutation distribution in \Eq~\ref{eq:permutation dist}.

\subsubsection{\bf Scenario 3: Order does not matter} This scenario is a special case of \textbf {Scenario 2} where the assignment (order) between the truth and the outputs can not be fixed during training step; However in this case, we do not consider learning the ordering structure (if exist any) in the annotations. 
For example, in the object detection problem, we may consider knowing the state of set of bounding boxes only, but ignoring how the bounding boxes are assigned (permuted) to the model output during training.

Therefore, in this case the distribution over the permutations, \ie $p_m(\bpi|\cdot,\cdot)$, can be assumed to be uniform, \ie a normalized constant value for all permutations and all input instances, 
\begin{equation}
\begin{aligned}
p_m(\bpi|\bx_{i},\bw) = p_m(\bpi) = \omega\qquad \forall \bpi\in\bPi 
%\\ = & \omega\qquad\forall\forall \bpi\in\bPi,
\label{eq:permutation dist-s3}
\end{aligned}
\end{equation}
   
 In this case, the posterior distribution in each SGD iteration (\Eq~\ref{eq:appr-posterior}) is simplified as \vspace{-0.5em} 
\begin{equation}
\begin{aligned}
p(\bw_k|\D) \!\!\!\!\!\!\!\appropto\!\!\!\!\!\!\prod_{i=1}^{n}\!&\Bigg[p(m_{i}|\bx_{i},\bw_k)\!\times 
\!U^{m_i}\!\!\times\!\omega \\&\quad\times\!\!\!\prod_{\sigma=\pi_1}^{\pi_{m_i}}p(\by_{\sigma}|\bx_{i},\bw_k,\bpi^*_{i,k}) \Bigg]p(\bw_k),
\label{eq:appr-posterior-s3}
\end{aligned}   
\end{equation} 
where $\bpi_{i,k}^*$ is the best permutation sample for each instance $i$ in each iteration of SGD $k$, attained by solving the following assignment problem:
\vspace{-0.5em}  
\begin{equation}
\label{eq:alternation1b}
\begin{aligned}
\bpi_{i,k}^*&=\arg\min_{\bpi\in\bPi}  -\log\bigg(\prod_{\sigma=\pi_1}^{\pi_{m_i}}p(\by_{\sigma}|\cdot,\cdot,\bpi)\bigg)\\&=\arg\min_{\bpi\in\bPi} \quad \sum_{\sigma=\pi_1}^{\pi_{m_i}}\calL_{state}\big(\by_{\sigma},\bO^{\sigma}_1(\bx_{i},\bw_{k-1})\big)\\
\end{aligned}
\vspace{-0.5em}
\end{equation}
To learn the parameters $\bw$, the sampled permutation $\bpi_{i,k}^*$ is used for back-propagation, \ie  
\vspace{-0.5em}  
\begin{equation}
\label{eq:alternation2b}
\begin{aligned}
\bw_{k}& = \bw_{k-1}-\eta\frac{-\partial \log\left(p\left(\bw_{k-1}|\D\right)\right)}{\partial\bw_{k-1}}\\
& = \bw_{k-1}-\eta\sum_{i=1}^{n}\Big[\frac{\partial \calL_{card}\big(m_i,\balpha\big)}{\partial\balpha}.\frac{\partial \balpha}{\partial\bw}+\\&\qquad\qquad\sum_{\sigma=\pi^*_1}^{\pi^*_{m_i}}\frac{\partial \calL_{state}\big(\by_{\sigma},\bO^{\sigma}_1\big)}{\partial\bO^{\sigma}_1}.\frac{\partial \bO^{\sigma}_1}{\partial\bw}\!\Big]+ 2\gamma\bw.
\end{aligned}
\end{equation}

\textbf{Implementation insight:} The definition and implementations details for the outputs, \ie $\balpha (\cdot,\cdot))$ and $\bO^{\sigma}_1 (\cdot,\cdot)$ and their losses, \ie $\calL_{card}(\cdot,\balpha (\cdot,\cdot))$, and $\calL_{state}(\cdot,\bO^{\sigma}_1 (\cdot,\cdot))$ are identical to \textbf{Scenario 2}. The only difference is that the term $\calL_{perm}(\cdot,\bO_2 (\cdot,\cdot))$ disappears from \Eqs~\ref{eq:alternation1} and~\ref{eq:alternation2}.

Closely looking into \Eqs~\ref{eq:alternation1}, ~\ref{eq:alternation2}, ~\ref{eq:alternation1b} and~\ref{eq:alternation2b} in both \textbf{Scenarios 2 and 3} and considering $\calL_{state}(\cdot,\cdot)$ only in these equations, this loss can be found under different names in the literature, \eg Hungarian~\cite{Stewart:2016:CVPR}, Earth mover's and Chamfer~\cite{fan2017point} loss, where an assignment problem between the output of the network and ground truth needs be determined \emph{before} the loss calculation and back-propagation. However, as explained in \Eq~\ref{eq:permutation dist}, these types of the losses are not actually permutation invariant losses and they are indeed a practical approximation for sampling (the best sample) from the permutation, as latent variable, in each SGD iteration.   

\subsection{Inference}
\label{sec:inference}
The inference process for all three scenarios is identical because the inference is independent from the definition of the permutation distribution as shown below. 

Having learned the network parameters $\bw^{*}$, for a test input $\bx^{+}$, we use a MAP estimate to generate a set output, \ie  $\calY^{*} = \displaystyle \arg\min_{\calY}\enspace-\log\left(p(\calY|\D,\bx^{+},\bw^{*})\right)$\vspace{-0.5em} 
%\begin{equation}
%\calY^{*} 
%= \arg\max_{\calY}\enspace p(\calY|\D,\bx^{+},\bw^{*}),
%\end{equation}
%where $p(\calY|\D,\bx^{+},\bw^{*})  \propto \int p(\calY|\bw,\bx^{+})p(\bw|\D) d\bw$, and $p(\bw|\D) = \delta(\bw=\bw^{*}|\D)$ as above. 
%Therefore, the MAP estimate can be written as follows, 
\begin{equation}
\begin{aligned}
\calY^{*} = &\arg\!\!\min_{m,\calY^m} \enspace\!\! -\log \left(p(m|\bx^{+},\bw^{*})\right)- m\log U-\\ &\log\!\!\sum_{\bpi\in\bPi}\left(p_m(\bpi|\bx^{+},\bw^{*})\times\!\!\!\prod_{\sigma=\pi_1}^{\pi_{m}}p(\by_{\sigma}|\bx^{+},\bw^{*},\bpi)\right).\notag
\vspace{-0.5em}
\end{aligned}
\label{eq:inference}
\end{equation}
Note that in contrast to the learning step, the process how the set elements during the prediction step are sorted and represented, does not affect the output values. Another way to explain this is that the permutation is defined during training procedure only as it is applied on the ground truth labels for calculating the loss. Therefore, it does not affect the inference process. To this end, the product $\prod_{\sigma=\pi_1}^{\pi_{m}}p(\by_{\sigma}|\bx^{+},\bw^{*},\bpi)$ is identical for any possible permutation, \ie $\forall\bpi\in\bPi$.  Therefore, it can be factorized from the summation, \ie\vspace{-0.5em}
\begin{equation}
\begin{aligned}
&\log\sum_{\bpi\in\bPi}\Big(p_m(\bpi|\bx^{+},\bw^{*})\times\prod_{\sigma=\pi_1}^{\pi_{m}}p(\by_{\sigma}|\bx^{+},\bw^{*},\bpi)\Big)\\
&=\log\Bigg(\prod_{\sigma=1}^{m}p(\by_{\sigma}|\bx^{+},\bw^{*})\times\underbrace{\sum_{\bpi\in\bPi}p_m(\bpi|\bx^{+},\bw^{*})}_{=1}\Bigg)\\
&=\sum_{\sigma=1}^{m}\log \left(p(\by_{\sigma}|\bx^{+},\bw^{*})\right)\notag.
\vspace{-0.5em}
\end{aligned}
\end{equation}
Therefore, the inference is simplified to 
\begin{equation}
\begin{aligned}
\calY^{*} 
= \arg\!\!\min_{m,\calY^{m}} \enspace -\log &\big( \underbrace{p(m|\bx^{+},\bw^{*})}_{\balpha}\big)- m\log U\ - \\&\sum_{\sigma=1}^{m}\log \big( \underbrace{p(\by_{\sigma}|\bx^{+},\bw^{*})}_{\bO^\sigma_1}\big).
\end{aligned}
\vspace{-0.5em}
\label{eq:inference2}
\end{equation}

\textbf{Implementation insight:} Solving \Eq~\ref{eq:inference2} corresponds to finding the optimal set $\calY^{*} = (m^*,\calY^{m^*})$, which is the \textbf{exact} solution of the problem.  As shown in~\cite{rezatofighi2018joint}, this problem can be optimally and efficiently calculated using few simple operations, \eg sorting, max and summation operations.
Note that the unit of hyper-volume $U$ is assumed as a constant hyper-parameter, tuned from the validation set of the data such that the best performance on this set is ensured.

An alternative to evaluate the models is an approximated inference~\cite{rezatofighi2017deepsetnet}, where the set cardinality is first approximated using $m^{*} 
= \arg\!\!\min_{m} \enspace -\log \big( p(m|\bx^{+},\bw^{*})\big)$ and then all $M$ set elements are sorted according to their existence scores and the state of top $m^{*}$ set elements are shown as the final output. The main key difference between the exact and approximate inferences is in the calculation of finding the optimal cardinality, where the approximate inference finds $m^{*}$ using the cardinality distribution only, while the exact inference considers all cardinality-related terms in \Eq~\ref{eq:inference2}, \ie the cardinality distribution, hyper-parameter $U$ and distribtion over the existence scores of the set elements in order to calculate $m^{*}$.         

\section{Experimental Results}
\label{sec:results}

%\fixmel{I feel there should be a diagram of the network here, so that people can see at a glance what a network for object detection would look like under a set formulation. Currently, there is no intuition presented in the paper about how one can actually design a network to do set learning.}
To validate our proposed set learning approach, we
perform experiments for each of the proposed scenarios including 
\emph{i)} Multi-label image classification (\textbf{Scenario 1}), 
\emph{ii)} object detection and identification on synthetic data (\textbf{Scenario 2}), 
\emph{iii)} 
(a) \revised{object} detection on three real datasets and 
(b) a CAPTCHA test to partition a digit into its summands  (\textbf{Scenario 3}).
% All are appropriate applications for our model
% as their annotated labels are expected to be in the form of a set, either a set of category tags, a set of bounding boxes or a set of candidate digits, with unknown cardinality and permutation.

\subsection{Multi-label image classification (Scenario 1)}\label{subsec:imagetagging}
To validate the first learning model in \Sec~\ref{sec:scenario1} (\textbf{Scenario 1}), we perform experiments on the task of multi-label image classification. This is a suitable application as the output is expected to be in the form of a set, \ie a set of category labels with an unknown cardinality while the order of its elements (category labels) in the output list does not have any preferable order. However, their order with respect to the network's output can be consistently fixed during training. Therefore, the best learning scheme for this application is $\textbf{Scenario 1}$, where the permutation is not a variable of the training process. To evaluate the framework, we use two standard public benchmarks for this problem, the PASCAL VOC 2007 dataset~\cite{Everingham:2007:PASCAL-VOC} and the Microsoft Common Objects in Context (MS COCO) dataset~\cite{lin2014microsoft}. 
%\fixmea{Image classification is immediately associated with ImageNet. We need to explain here, even if very briefly, why we do not present results on ImageNet.}

\myparagraph{Implementation details.} 
To train a network for predicting a set of tags for an image, we use the training scheme detailed as \textbf{Scenario~1}. To ensure a fair comparison with other works, we adopt the same base network architecture as~\cite{gong2013deep,wang2016cnn,rezatofighi2017deepsetnet}, \ie $16$-layers VGG network~\cite{Simonyan:2014:VGG} pretrained on the 2012 ImageNet dataset, as the backbone of our set prediction network. We adapt VGG for our purpose by modifying the last fully connected prediction layer to predict both cardinality $\balpha(\cdot,\cdot)$ and the state of $M$ set elements $\bO_1 = (\bO^1_1,\bO^2_1,\cdots,\bO^M_1)$, which are only the existence scores for all $M$ category labels in this problem. To train the model using \Eq~\ref{eq:map_complete_S1}, we define categorical distribution for cardinality and use softmax (Cross Entropy) loss as $\calL_{card}$ and also binary cross-entropy (BCE) as $\calL_{card}$. We then fine-tune the entire network using the training set of these datasets with the same train/test split as in existing literature~\cite{gong2013deep,wang2016cnn}. 

\begin{figure}[tb]
	\vspace{-1em}
% 	\begin{minipage}[b]{.325\linewidth}
% 		\includegraphics[width=\linewidth]{figs/F1_vs_U.png}
% 		\centerline{(a)}\medskip
% 	\end{minipage}
% 	\hfill
	\begin{minipage}[b]{.48\linewidth}
		\includegraphics[width=\linewidth]{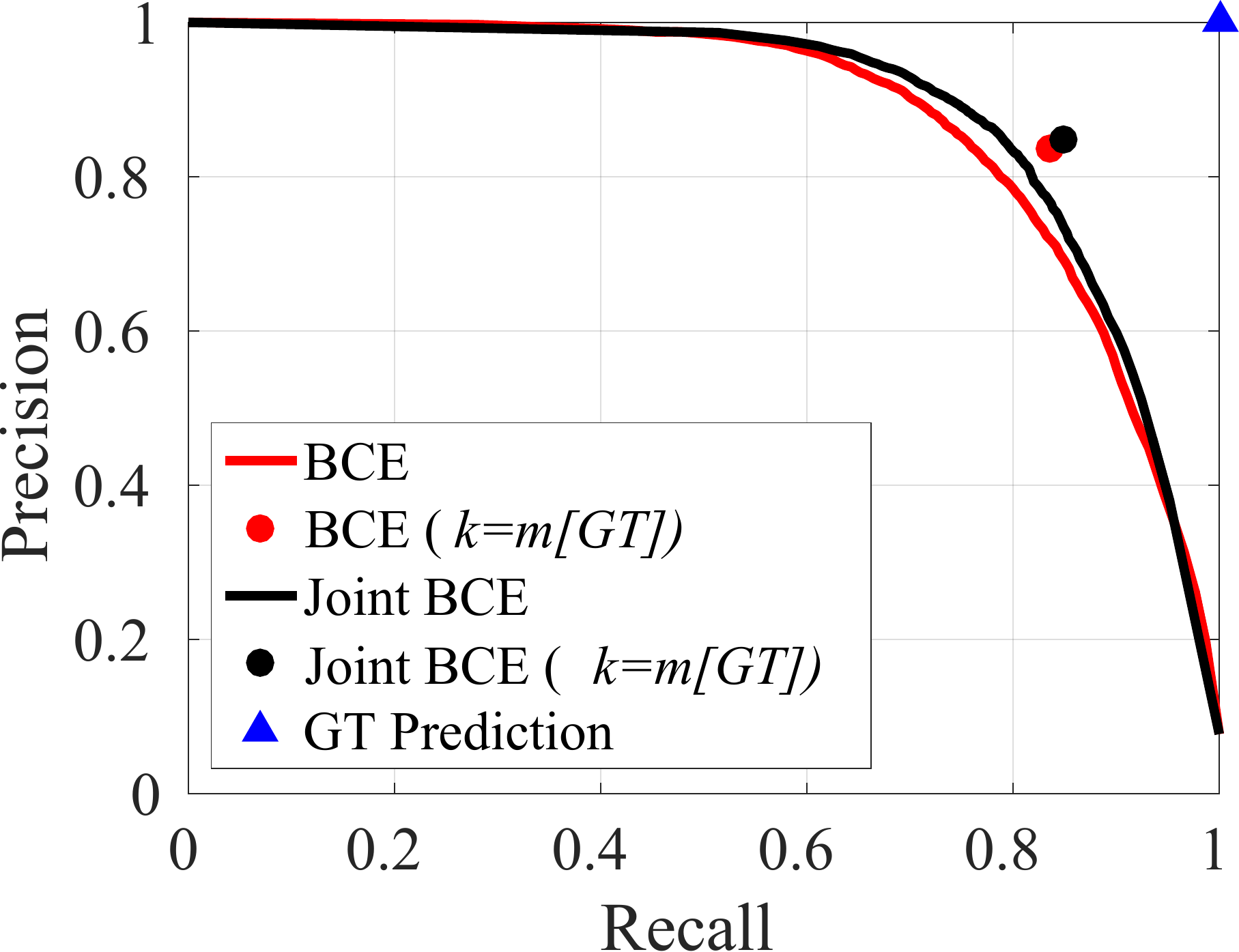}
		\centerline{(a)}\medskip
	\end{minipage}
		\hfill
		\begin{minipage}[b]{.55\linewidth}
		\includegraphics[width=\linewidth]{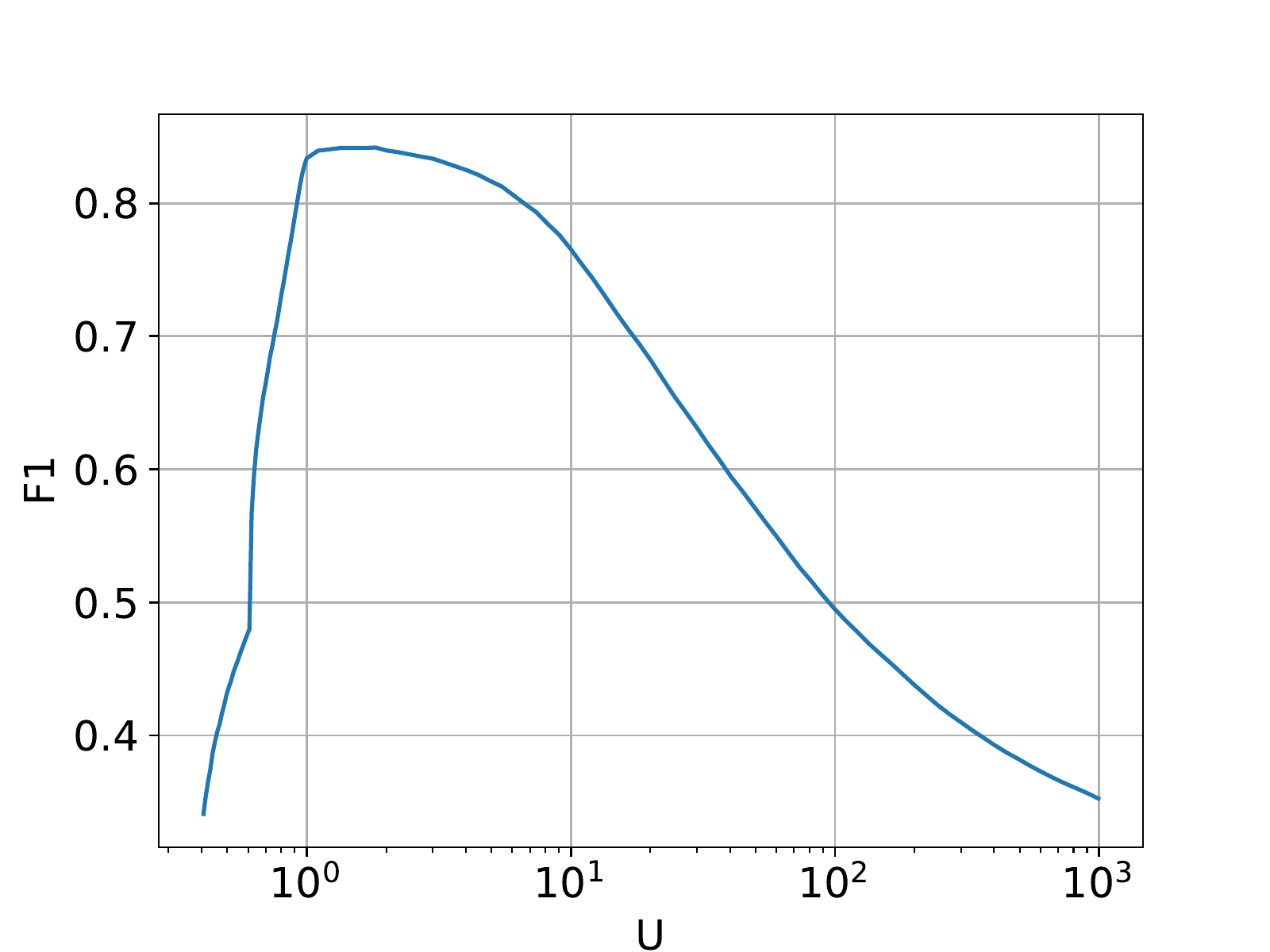}
		\centerline{(b)}\medskip
	\end{minipage}

	\vspace{-1.0em}
	\small
		\caption{\revised{(a) Precision/recall curves for the classification scores when the classifier is trained independently (red solid line) and when it is trained jointly with the cardinality term using our proposed joint approach (black solid line) on PASCAL VOC dataset. The circles represent the upper bound when ground truth cardinality is used for the evaluation of the corresponding  classifiers. The ground truth prediction is shown by a blue triangle. (b) The value of F1 against the value of hyper-parameter $U$ in log-scale (PASCAL VOC). }}\label{fig:curves-mlic}

	\label{fig:U_param}
\end{figure}

To optimize \Eq~\ref{eq:map_complete_S1}, we use stochastic gradient descent (SGD) and set the weight decay to $\gamma = 5\cdot 10^{-4}$, with a momentum of $0.9$ and a dropout rate of $0.5$.  The learning rate is adjusted to gradually decrease after each epoch, starting from $0.001$. The network is trained for $60$ epochs for both datasets and the epoch with the lowest validation objective value is chosen for evaluation on the test set. The hyper-parameter $U$ is set to be $2.36$, adjusted on the validation set. 
\revised{Note that a proper tuning of $U$ can be important for the model's superior performance when the exact inference is applied (as shown in \Fig~\ref{fig:curves-mlic} (b)). Alternatively, the approximate inference which does not rely on $U$ can be applied.}

In our earlier work~\cite{rezatofighi2017deepsetnet}, we proposed the first set model based on \textbf{Scenario 1} only. However, we used two separate networks (two $16$-layers VGG networks), one to model the set state (class scores) and one for cardinality. We also used the negative binomial as a  cardinality loss. In addition, the trained model used approximate inference to generate the final set at test time. Here, we compare our proposed framework with the model from~\cite{rezatofighi2017deepsetnet} to demonstrate the advantage of joint learning for each of the optimizing objectives, \ie label's existence scores and their cardinality distribution. To ensure a fair comparison with~\cite{rezatofighi2017deepsetnet}, we add another baseline experiment where we use the same cardinality model but replace the negative binomial (NB) distribution used in~\cite{rezatofighi2017deepsetnet} with the categorical distribution (Softmax). We follow the exact the same training protocol as~\cite{rezatofighi2017deepsetnet} to train the classifier and cardinality network for the baseline approach.  

\myparagraph{Evaluation protocol.}
% \subsubsection{Evaluation protocol}
We employ the common evaluation metrics for multi-label image classification also used in~\cite{gong2013deep,wang2016cnn}. These include the average \textit{precision}, \textit{recall} and \textit{F1-score}\footnote{F1-score is calculated as the harmonic mean of precision and recall.} of the generated labels, calculated per-class (C-P,  C-R and C-F1) and overall (O-P,  O-R and O-F1). Since C-P,  C-R and C-F1 can be biased to the performance of the most frequent classes, we also report the average \textit{precision}, \textit{recall} and \textit{F1-score} of the generated labels per image/instance (I-P, I-R and I-F1). 

% Since precision and recall can be different for each approach and cannot be a proper representative metric, we rely on F1 score to rank approaches on the final task of label prediction. The perfect set prediction will have F1 score equal to 1, which is a single point in the top right corner of the precision and recall plot (see blue triangle in Fig.~\ref{fig:curves-mlic}). 
We rely on F1-score to rank approaches on the task of label prediction. A method with better performance has a precision/recall value that has a closer proximity to the perfect point shown by the blue triangle in Fig.~\ref{fig:curves-mlic}. % This property is well reflected in the F1 score. 
%  Precision is defined as the ratio of
%correctly predicted labels and total predicted labels, while
%recall is the ratio of correctly predicted labels and ground-truth labels. 
%In case no predictions (or ground truth) labels exist, \ie the denominator becomes zero, precision (or recall) is defined as $\%100$. 
To this end, for the classifiers such as BCE and Softmax, we find the optimal evaluation parameter $k=k^*$ that maximises the F1-score. For the set network model in~\cite{rezatofighi2017deepsetnet} (DS) and our proposed framework with joint backbone network (JDS), prediction/recall is not dependent on the value of $k$. Rather, one single value for precision, recall and F1-score is computed. 
\\\\
% \subsubsection{PASCAL VOC 2007}
\myparagraph{PASCAL VOC 2007.} 
% \begin{figure}[t]
% \centering
% 	\includegraphics[width=0.7\linewidth]{figs/VOC_Curve_JDS2.pdf}
% 	\caption{Precision/recall curves for the classification scores when the classifier is trained independently (red solid line) and when it is trained jointly with the cardinality term using our proposed joint approach (black solid line) on PASCAL VOC dataset. The circles represent the upper bound when ground truth cardinality is used for the evaluation of the corresponding  classifiers. The ground truth prediction is shown by a blue triangle.}\label{fig:curves-mlic}
% \end{figure}
%

%

\newcommand{\colw}{0.43cm}
\newcommand{\colww}{0.53cm}
\newcommand{\lsh}{\!\!\!\!}
\begin{table*}[tbh]
\revised{
\footnotesize
	\caption{Quantitative results for multi-label image classification on (a) the PASCAL VOC and (b) the MS COCO datasets.}
	\vspace{-1em}
	\begin{center}
		\begin{tabular}{lcc||p{\colw}p{\colw}p{\colww}| p{\colw}p{\colw}p{\colww}|p{\colw}p{\colw}p{\colww} @{}}
			% 			\raisebox{-1.0ex}{Classifier}& \raisebox{-1.0ex}{/Metric} & \raisebox{-1.0ex}{C-P} & \raisebox{-1.0ex}{C-R} & \raisebox{-1.0ex}{C-F1} & \raisebox{-1.0ex}{O-P} & \raisebox{-1.0ex}{O-R} & \raisebox{-1.0ex}{O-F1}\\
			% 			&\raisebox{-0.5ex}{Evaluation}&&&&&&\\
			\multicolumn{11}{c}{(a) PASCAL VOC}\\
			\lsh Output-Method (Loss) &Backbone/Decoder& Eval. & \scriptsize{C-P} & \scriptsize{C-R} & \scriptsize{C-F1} & 
			\scriptsize{O-P} & \scriptsize{O-R} & \scriptsize{O-F1} & \scriptsize{I-P} & \scriptsize{I-R} & \scriptsize{I-F1} \\
			\hline\hline % inserts single-line
			% Entering 1st row
			%			Upper Band&k=3&$64.1$&$74.4$&$68.9$&$75.1$&$78.3$&$76.7$\\ 
			%			\hline
			\lsh Tensor-B/L (Softmax)&Full VGG16&k=$k^*$(1)&$88.2$&$65.4$&$75.1$ &$91.3$&$59.2$&$71.8$&$91.3$&$69.8$&$79.1$\\
			%			WARP&k=3&&&&&&\\
			\lsh Tensor-B/L (BCE)&Full VGG16&k=$k^*$(1)&$88.7$&$58.6$&$70.5$&$92.2$&$59.8$&$72.5$&$92.2$&$70.1$&$79.6$\\
			\hline
			\lsh \bf Set-DS (BCE-NB)~\cite{rezatofighi2017deepsetnet}&Full VGG16&k=$m^*$ &$76.8$&$74.8$&$75.8$&$80.6$&$76.7$&$78.6$&$83.4$&$81.9$&$82.6$\\
			%			\lsh DS (BCE-NB)~\cite{}&k=$m^{GT}$ &$78.8$&$80.8$&$79.8$&$83.5$&$83.5$&$83.5$&$85.7$&$85.7$  &$85.7$\\
			\lsh \bf Set-DS (BCE-SftMx)&Full VGG16&k=$m^*$&$77.1$ & $75.2$ &$76.2$ & $81.0$ &  $77.1$ & $79.0$ & $83.9$ & $82.1$ & $83.0$\\
			%\lsh\textbf{Set-JDS (BCE-SftMx)}&Full VGG16&k=$m^*$ &$83.5$&$74.4$&$\textbf{78.7}$&$85.5$&$77.9$&$\textbf{81.5}$&$87.6$&$82.8$&$\textbf{85.1}$\\
			\lsh\textbf{Set-JDS (BCE-SftMx)}&Full VGG16&k=$m^*$ &$86.3$&$78.5$&$\textbf{82.2}$&$88.6$&$80.3$&$\textbf{84.2}$&$95.2$&$89.9$&$\textbf{92.5}$\\
			\multicolumn{11}{c}{}\\
			\multicolumn{11}{c}{(b) MS COCO}\\
						\hline\hline % inserts single-line
			% Entering 1st row
			%			Upper Band&k=3&$64.1$&$74.4$&$68.9$&$75.1$&$78.3$&$76.7$\\ 
			%			\hline
			\lsh Tensor-B/L (Softmax)&Full VGG16&k=$k^*$(3)&$58.6$&$57.6$&$58.1$ &$60.7$&$63.3$&$62.0$&$60.7$&$74.7$&$67.0$\\
			%			WARP&k=3&&&&&&\\
			\lsh Tensor-B/L (BCE) &Full VGG16&k=$k^*$(3)&$56.2$&$60.1$&$58.1$&$61.6$&$64.2$&$62.9$&$61.6$&$75.3$&$67.8$\\
			\hline
			\lsh Sequential-CNN+RNN~\cite{wang2016cnn} (BCE) &Full VGG16&k=$k^*$(3)&$66.0$&$55.6$&$60.4$&$69.2$&$66.4$&$67.8$&$-$&$-$&$-$\\
			\hline
			\lsh \bf Set-DS (Softmax-NB)~\cite{rezatofighi2017deepsetnet}&Full VGG16&k=$m^*$ &$68.2$&$59.9$&$63.8$&$68.8$&$67.4$&$68.1$&$74.3$ & $72.6$ & $73.5$\\
			%			\textbf{DeepSetNet (WARP)}&Est. Card.&&&&&&\\
			\lsh \bf Set-DS (BCE-NB)~\cite{rezatofighi2017deepsetnet}&Full VGG16&k=$m^*$ &$66.5$&$62.9$&$64.6$&$70.1$&$68.7$&$69.4$&$75.2$&$73.6$&$74.4$\\	
			%			WARP&k=3&&&&&&\\
			\lsh \bf Set-DS (BCE-Sftmx)&Full VGG16& k=$m^*$& $68.0$ & $61.7$ & $64.7$ & $72.4$ & $67.1$ & $69.6$& $76.0$ & $73.3$ & $74.6$\\
			\lsh\textbf{Set-JDS (BCE-Sftmx)}&Full VGG16&k=$m^*$ &$70.2$&$61.5$&$\textbf{65.5}$&$74.0$&$67.6$&$\textbf{70.7}$&$77.9$&$73.4$&$\textbf{75.6}$\\
		\end{tabular}
	\end{center}
	\label{table:allvoc-coco-multilabel}}
%	\vspace{-.5em}
\end{table*}

We first test our approach on the Pascal Visual Object Classes benchmark~\cite{Everingham:2007:PASCAL-VOC}, which is one of the most widely used datasets for detection and classification. This dataset includes $9963$ images with a 50/50 split for training and test, where objects from $20$ pre-defined categories have been annotated by bounding boxes. Each image  contains between $1$ and $7$ unique objects. 

% \begin{table*}[tbh]
% 	\caption{Quantitative results for multi-label image classification on the MS COCO dataset.}
% %	\vspace{-1em}
% 	\begin{center}
% 		\begin{tabular}{lc||p{\colw}p{\colw}p{\colww}| p{\colw}p{\colw}p{\colww}|p{\colw}p{\colw}p{\colww} @{}}
% 			% 			\raisebox{-1.0ex}{Classifier}& \raisebox{-1.0ex}{/Metric} & \raisebox{-1.0ex}{C-P} & \raisebox{-1.0ex}{C-R} & \raisebox{-1.0ex}{C-F1} & \raisebox{-1.0ex}{O-P} & \raisebox{-1.0ex}{O-R} & \raisebox{-1.0ex}{O-F1}\\
% 			% 			&\raisebox{-0.5ex}{Evaluation}&&&&&&\\
% 			\lsh Classifier & Eval. & \scriptsize{C-P} & \scriptsize{C-R} & \scriptsize{C-F1} & 
% 			\scriptsize{O-P} & \scriptsize{O-R} & \scriptsize{O-F1} & \scriptsize{I-P} & \scriptsize{I-R} & \scriptsize{I-F1} \\

% 			%			\hline
% 			%			\lsh\textbf{JDS (BCE-DC)}&k=$m^{GT}$ &&&&&&&&&\\
% 			%			\lsh\textbf{JDS (Rnk-DC)}&k=$m^*$&&&&&&&&&\\
% 		\end{tabular}
% 	\end{center}
% 	\label{table:allcoco-multilabel}
% %	\vspace{-1.5em}
% \end{table*}

We first investigate if the learning using the shared backbone (joint learning) improves the performance of cardinality and classifier. Fig.~\ref{fig:curves-mlic} (a) shows the precision/recall curves for the classification scores when the classifier is trained solely using binary cross-entropy (BCE) loss (red solid line) and when it is trained using the same loss jointly with the cardinality term (Joint BCE). We also evaluate the precision/recall values when the ground truth cardinality $m[GT]$ is provided. The results confirm our claim that the joint learning indeed improves the classification performance. We also calculate the mean absolute error of the cardinality estimation when the cardinality term using the DC loss is learned jointly and independently as in~\cite{rezatofighi2017deepsetnet}. The mean absolute cardinality error of our prediction on PASCAL VOC is $0.31\pm0.54$, while this error is $0.33\pm0.53$ when the cardinality is learned independently. 

We compare the performance of our proposed set network, \ie JDS (BCE-DC), with traditional learning approaches trained using classification losses only, \ie softmax and BCE losses, while during the inference, the labels with the best $k$ value are extracted as \revised{the model's output} labels. 

For the set based approaches, we included our model in~\cite{rezatofighi2017deepsetnet} when the classifier is binary cross entropy and the cardinality loss is negative binomial, \ie DS (BCE-NB). In addition, Table~\ref{table:allvoc-coco-multilabel} (a) reports the results for the deep set network when the cardinality loss is replaced by a categorical loss, \ie (BCE-SftMx). The results show that we outperform the other approaches \wrt all three types of F1-scores. In addition, our joint formulation allows for a single training step to obtain the final model, while our model in ~\cite{rezatofighi2017deepsetnet} learns two VGG networks to generate the output sets.  

\myparagraph{Microsoft COCO. } 
\begin{figure*}[t]
\centering
	\includegraphics[width=0.9\linewidth]{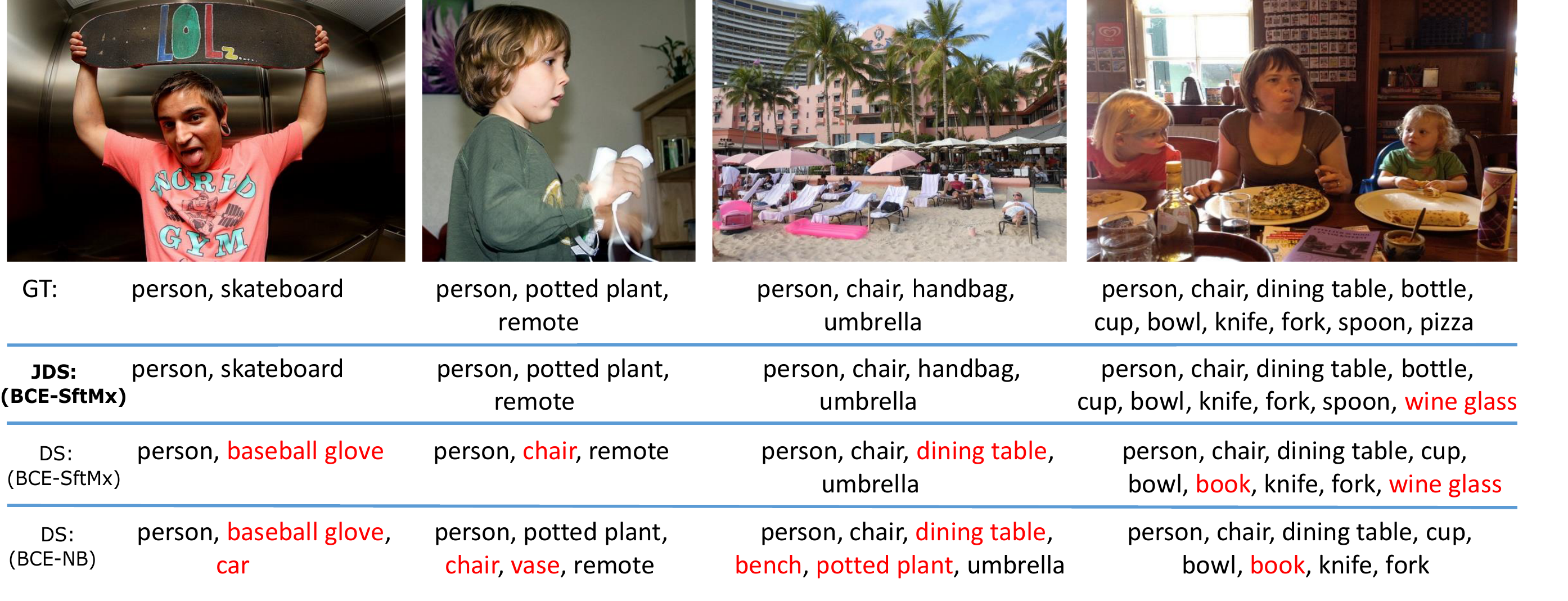}
	\caption{Qualitative comparison between our proposed set network with a shared backbone (JDS) and the deep set networks with softmax (DS (BCE-Sftmx)) and Negative Binomial (DS (BCE-NB)) as the cardinality loss. For each image, the ground truth tags and the predictions for our JDS and the two baselines are denoted below. {\textcolor{red}{False positives}} are highlighted in red. Our JDS approach reduces both cardinality and classification error.} 
%	\vspace{-1.0em}
	\label{fig:Results1}
\end{figure*}

% % \fixmea{Report cardinality estimation error?}\fixmeh{mean absolute error = $0.3241$ and std absolute error  = $0.5235$ on the test set}

%  \begin{figure*}[t]
%  	\centering
%  	\includegraphics[width=.49\linewidth]{figs/PerClass_ROC_Curve_VOC.pdf}
%  	\includegraphics[width=.49\linewidth]{figs/Overal_ROC_Curve_VOC.pdf}
%  	\caption{Pascal VOC 2007.} 
%  	\label{fig:curvesVOC}
%  	\end{figure*}

%
The MS-COCO~\cite{lin2014microsoft} benchmark is another popular benchmark for image captioning, recognition, and segmentation. The dataset includes
$123$K  images and over $500k$ annotated object instances from $80$ categories. The number of unique objects for each image  varies between $0$ and $18$. Around $700$ images in the training set do not contain any of the $80$ classes and there are only a handful of images that have more than $10$ tags. Most images contain between one and three labels.  We use $83$K
images with identical training-validation split as~\cite{rezatofighi2017deepsetnet}, and the remaining $40$K images as test data. 
%We predict the cardinality of objects in the scene with a mean absolute error of $0.74\pm0.86$.
%%
%%\input{tables/allcoco-multilabel}
%
%
%\Fig~\ref{fig:curves-mlic}(b) shows a significant improvement of precision and recall and consequently the F1 score using our deep set network compared to the softmax and binary cross-entropy classifiers for all ranking values $k$. 

The classification results on this dataset are reported in Tab.~\ref{table:allvoc-coco-multilabel} (b). The results once again show that our approach consistently outperforms our traditional and set based baselines measured by F1-score. Our approach also outperform a recurrent based approach~\cite{wang2016cnn}, which uses VGG-16 as the backbone along with a recurrent neural network to generate a set of labels. Due to this improvement, we achieve  state-of-the-art results on this dataset as well. 
Some examples of label prediction using our proposed set network model and comparison with other deep set networks are shown in \Fig~\ref{fig:Results1}. 
%The results show that our joint learning can simultaneously reduce the cardinality and classification errors in these examples. 

\subsection{Object detection (Scenario 2 \& 3)}
\label{result:od}
\newcommand{\shft}{\hspace{-.2cm}}
\revised{Our second experiment is used to test our set formulation for the task of object detection using the model trained according to \textbf{Scenario 3} on \revised{three real detection dataset, including \emph{(i)} a processed and cropped pedestrian dataset from MOTChallenge~\cite{mot2015challenge, mot17detchallenge}, \emph{(ii)} PASCAL VOC 2007~\cite{Everingham:2007:PASCAL-VOC}, and \emph{(iii)} MS COCO~\cite{lin2014microsoft}}. We also assess our set formulation for the task of object detection using the model trained according to \textbf{Scenario 2} on synthetically generated dataset in order to demonstrate the advantage and the limitation of this scenario. This experiment and all the details can be found in the supplementary material document.}

%We compare it with the popular object detectors, \ie \revised{Faster R-CNN}~\cite{ren2015faster} and YOLO~\cite{redmon2017yolo9000,redmon2018yolov3}. To ensure a fair comparison, we use the exact same base network structure (ResNet-101) and the same training dataset. Moreover, the best hyper parameters, \eg NMS threshold, are tuned for \revised{Faster R-CNN} and YOLO detectors. 

\revised{\textbf{Formulation, training losses and inference details.}
We formulate object detection as a set prediction problem $\calY=\left\{\by_1,\cdots,\by_m\right\}$, where each set element represents a bounding box as $\by = (x,y,w,h,\ell)\in \mathbb{R}^4\times\mathbb{L}$, where $(x, y)$ and
$(w, h)$ are respectively the bounding boxes’ position and size and $\ell\in\mathbb{L}$ represents the class label for this set element.}

\revised{We train a feed-forward neural network including a backbone and few decoder layers
using {\bf Scenario 3} with loss heads directly attached to the final output layer. According to Eqs.~(\ref{eq:alternation1b}, \ref{eq:alternation2b}), there are two main terms (losses) that need to be defined for this task. Firstly, for cardinality $\calL_{card}(\cdot)$, a categorical loss (Softmax) is used as the discrete distribution model. Secondly, the state loss $\calL_{state}(\cdot)$ consists of two parts in this problem: \emph{i)} the bounding box regression loss between the predicted output states and the permuted ground truth states, \emph{ii)} the classification loss for $\ell$. 
For the bounding box regression, we use two options:  1) Smooth L1-loss only and 2) Smooth L1-loss along with the Generalized Intersection over Union (GIoU)~\cite{rezatofighi2019generalized} loss. The cross-entropy loss is also used for the class label $\ell$. The permutation is estimated iteratively using alternation according to Eq.~(\ref{eq:alternation1b}) using Hungarian (Munkres) algorithm. }

\revised{For the pedestrian detection as a single class object detection problem, we use the exact inference, explained in \Sec\ref{sec:inference}. However, since the extension of this exact inference for multi-class object detection is not trivial, we apply the approximated inference (\Sec\ref{sec:inference}) by estimating the total number of objects using the highest value in the cardinality Softmax, $m^*$ and extracting $m^*$ top bounding boxes with the highest Softmax class scores (independent from their class labels).}

\revised{\textbf{Ablation and Comparison.} To demonstrate the effect of network backbone and decoder layers in the final results, we incorporate various state-of-the art feed-forward architectures in our experiments including \emph{i)}  standard Full ResNet-101 architecture~\cite{he2016deep} including its convolutional layers as the backbone and its FC layers as the decoder, \emph{ii)} ResNet-50 base~\cite{he2016deep} with/without the transformer encoder architecture used in~\cite{vaswani2017attention} as the backbone and the transformer encoder/decoder~\cite{vaswani2017attention} as the decoder part of the model. Note that our proposed model using ResNet-50 with the transformer encoder and decoder without the cardinality term would be identical to DETR~\cite{carion2020end}, one of the state-of-the art object detection frameworks.}
\revised{We also compare our set-based detector baseline models with two popular tensor/anchor-based object detectors, \ie \revised{Faster R-CNN}~\cite{ren2015faster} and YOLO v3~\cite{redmon2018yolov3} as well as a sequential anchor-free model~\cite{Stewart:2016:CVPR}.} 

\revised{\textbf{Evaluation protocol.} To quantify the detection performance on MOTChallenge and PASCAL VOC datasets, we adopt the
commonly used metrics~\cite{Dollar:2012:PAMI} such as mean average precision (mAP) and the log-average miss rate (MR) over false positive per image, using the metric parameter $IoU = 0.5$. For the experiments on the COCO datasets, we use COCO evaluation metric, \ie Average Precision (AP), attained by averaging mAP over different values of $IoU$ between $0.5$ and $0.95$.  
Additionally, we report the F1 score (the
harmonic mean of precision and recall) for our model (when evaluated using cardinality during inference) and for all the other baseline methods using their best confidence threshold.  In order to reflect the computation and the complexity of the used feed-forward network architectures, their total number of learnable parameters and Flops are also reported. }

\subsubsection{\revised{Pedestrian detection (Small-scale MOTChallenge)}}
\begin{figure*}[t]
	\begin{minipage}[b]{.32\linewidth}
		\begin{minipage}[b]{.49\linewidth}
			\includegraphics[width=.99\linewidth]{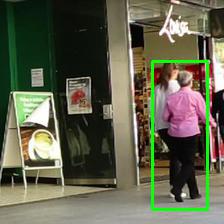}
		\end{minipage}
		\begin{minipage}[b]{.49\linewidth}
			\includegraphics[width=.99\linewidth]{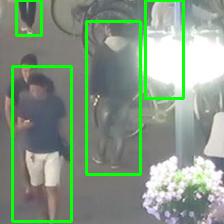}
		\end{minipage}
		\centerline{(a)}\medskip
	\end{minipage}
	\hfill
	\begin{minipage}[b]{.32\linewidth}
		\begin{minipage}[b]{.49\linewidth}
			\includegraphics[width=.99\linewidth]{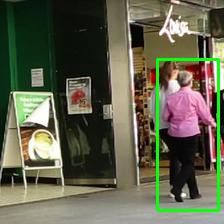}
		\end{minipage}
		\begin{minipage}[b]{.49\linewidth}
			\includegraphics[width=.99\linewidth]{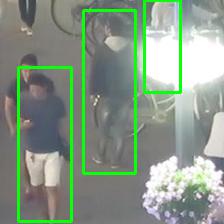}
		\end{minipage}
		\centerline{(b)}\medskip
	\end{minipage}
	\hfill
	\begin{minipage}[b]{.32\linewidth}
		\begin{minipage}[b]{.49\linewidth}
			\includegraphics[width=.99\linewidth]{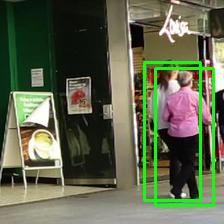}
		\end{minipage}
		\begin{minipage}[b]{.49\linewidth}
			\includegraphics[width=.99\linewidth]{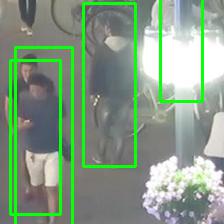}
		\end{minipage}
		\centerline{(c)}\medskip
	\end{minipage}
	\vspace{-1.0em}
	
	%		\includegraphics[width=.31\linewidth]{figs/frcnn3.jpg}
	%			\includegraphics[width=.31\linewidth]{figs/SND1.jpg}
	%			\includegraphics[width=.31\linewidth]{figs/SND2.jpg}
	%			\includegraphics[width=.31\linewidth]{figs/SND3.jpg}
	% 'MOT16-RegProps', 'MOT16-ours', 'MOT16-MS-CNN'
	\small
	\caption{ \small A qualitative comparison between the detection performance of (a) \revised{Faster R-CNN}, (b) YOLO v3 and (c) our basic set detection baseline (\ie ResNet101+L1-smooth loss) on heavily overlapping pedestrians from \emph{MOTChallenge} benchmark. Both \revised{Faster R-CNN} and YOLO v3 fail to properly detect heavily occluded pedestrians due to NMS. }
	\label{fig:Real_detections}
	\vspace{-1.em}
\end{figure*}
\textbf{Dataset.} We use training sequences from \emph{MOTChallenge} pedestrian detection and tracking benchmark, \ie 2DMOT2015~\cite{mot2015challenge} and MOT17Det~\cite{mot17detchallenge}, to create our dataset. Each sequence is split into train and test sub-sequences, and the images are cropped on different scales so that each crop contains up to 5 pedestrians. %Besides, we ensured that the cropped images have uniform level occlusions. 
 \revised{Our aim is to show the main weakness of tensor (anchor)-based object detectors such as Faster R-CNN~\cite{ren2015faster} and YOLO~\cite{redmon2017yolo9000,redmon2018yolov3}, relying on the NMS heuristics for handling partial occlusions, which is crucial in problems like multiple object tracking and instance level segmentation.} To this end, we first evaluate the approaches on small-scale data (up to 5 objects) which include a single type of objects, \ie pedestrian, with high level of occlusions. The resulting dataset has $50$K training and $5$K test samples.
%\fixmea{Why is this not evaluated properly on the official test set?}
\begin{table*}[bt]
	\footnotesize
	\caption{\small \revised{Detection results on the pedestrian detection dataset (Cropped MOTChallenge) measured by mAP, F1 score and MR rate. The abbreviations``BB", ``Card.'', ``Hun.'', ``Trans.'', ``En.'', and ``Dec.'' mean ``bounding box'', ``Cardinality'', ``Hungarian Algorithm'', ``Transformer'', ``Encoder'' and ``Decoder'', respectively. } }
	\vspace{-1.5em}
	\begin{center}
		\begin{tabular}{l|cccccccc}
			\toprule
			\shft \revised{Output (Method)} &\revised{Backbone$\slash$Decoder}& \revised{BB loss} & \revised{Card.} & mAP $\uparrow$  & F1$\uparrow$ & MR $\downarrow$ & \revised{Parameters} & \revised{Flops}\\
			\midrule
			\shft  \revised{Tensor (Faster R-CNN ~\cite{ren2015faster})}   &\revised{Full ResNet101}&\revised{L1}&\revised{\xmark}&$0.68$ & $0.76$ & $0.48$&\revised{42M}&\revised{23.2G}\\
			\shft  \revised{Tensor (YOLO v3~\cite{redmon2018yolov3}) }  &\revised{Full ResNet101}&\revised{L1}&\revised{\xmark}&$0.70$ & $0.76$ & $0.48$&\revised{59M}&\revised{24.4G}\\
			\hline
			            \shft   \revised{Sequential (LSTM-Hun.~\cite{Stewart:2016:CVPR})} &\revised{ResNet101$\slash$LSTM} & \revised{L1}        &\revised{\xmark}& \revised{$0.61$} & \revised{$0.62$} &\revised{$0.59$}&\revised{47.3M}&\revised{15.7G}\\
            
 			\hline
 			\shft \bf	\revised{Set (Ours)}	&\revised{Full ResNet101}& \revised{L1}&\revised{$-$}&\revised{$0.85$}& \revised{$0.86$} &\revised{$0.25$} &\revised{46.8M}&\revised{15.7G}\\
             \shft   \bf \revised{Set (Ours)}    &\revised{Full ResNet101}& \revised{L1+GIoU}      &\revised{$-$}& \revised{$0.85$} & \revised{$0.86$} &\revised{$0.24$}&\revised{46.8M}&\revised{15.7G}\\
             \shft  \bf \revised{Set (Ours)}    &\revised{Full ResNet101} & \revised{L1+GIoU}  &\revised{\checkmark} & \revised{$-$} & \revised{$0.86$} &\revised{$-$}&\revised{46.8M}&\revised{15.7G}\\
              \shft \bf \revised{Set (DETR~\cite{carion2020end})}    &\revised{ResNet50+Trans. En$\slash$Trans. Dec.} &\revised{ L1+GIoU}  &\revised{$-$}& \revised{$\bf{0.87}$} & \revised{$0.86$} & \revised{$\bf{0.17}$}&\revised{41M}&\revised{11.1G}\\
             \shft    \bf \revised{Set (Ours*)}    &\revised{ResNet50+Trans. En.$\slash$Trans. Dec.}  &\revised{L1+GIoU} & \revised{\checkmark}& \revised{$-$} & \revised{$\bf{0.87}$} & \revised{$-$}&\revised{41M}&\revised{11.1G}\\
			\bottomrule
		\end{tabular}
	\end{center}
	\label{table:cropmot new}
\end{table*}

\textbf{Detection results.} \revised{Quantitative detection results for all our set-based baselines as well as the other competing methods are shown in Tab.~\ref{table:cropmot new}. Since our framework generates a single set only (a set of bounding boxes) using the inference introduced in Sec.~\ref{sec:inference}, there exists one single value for
precision-recall and thus F1-score. For this reason, the average precision (AP) and log-average miss rate (MR) calculated over different thresholds cannot be reported in this case. To this end, we report these values on our set-based baselines using the predicted boxes with their scores only, and ignore the cardinality term and the inference step. These results are shown by``$-$'' below the column ``card.'' in Tab.~\ref{table:cropmot new}.} 
%To ensure a fair comparison, the F1-score reported in this table reflects the best score for \revised{Faster R-CNN} and YOLO v3 along the precision-recall curve.

\revised{Quantitative results in Tab.~\ref{table:cropmot new} demonstrates that our basic set-based detection baseline significantly outperforms, on all metrics, the tensor/anchor-based~\cite{ren2015faster,redmon2018yolov3} and the LSTM-based~\cite{Stewart:2016:CVPR} object detectors with an identical network architecture (ResNet101) and the same bounding box regression (L1-smooth) loss.}

\revised{We further investigate failure cases for the tensor-based~\cite{ren2015faster,redmon2018yolov3} object detectors, \ie \revised{Faster R-CNN} and YOLO v3, in Fig.~\ref{fig:Real_detections}. In case of heavy occlusions, the conventional formulation of these methods, which include the NMS heuristics, is not capable of correctly detecting all objects, \ie pedestrians.  Note that lowering the overlapping threshold in NMS in order to tolerate a higher level of occlusion results in more false positives for each object. In contrast, more occluding objects are miss-detected by increasing the value of this threshold. Therefore, changing the overlap threshold for NMS heuristics would not be conducive for improving their detection performances.
In contrast, our set learning formulation naturally handles heavy occlusions (see Fig.~\ref{fig:Real_detections}) by outputting a set of detections with no heuristic involved. Although, the LSTM-based object detector~\cite{Stewart:2016:CVPR} also does not rely on this heuristic to generate its final output, its formulation for object detection as a sequence prediction problem is a sub-optimal solution, resulting in an inferior performance.}

\begin{figure}[tb]
\centering
	\includegraphics[width=0.6\linewidth]{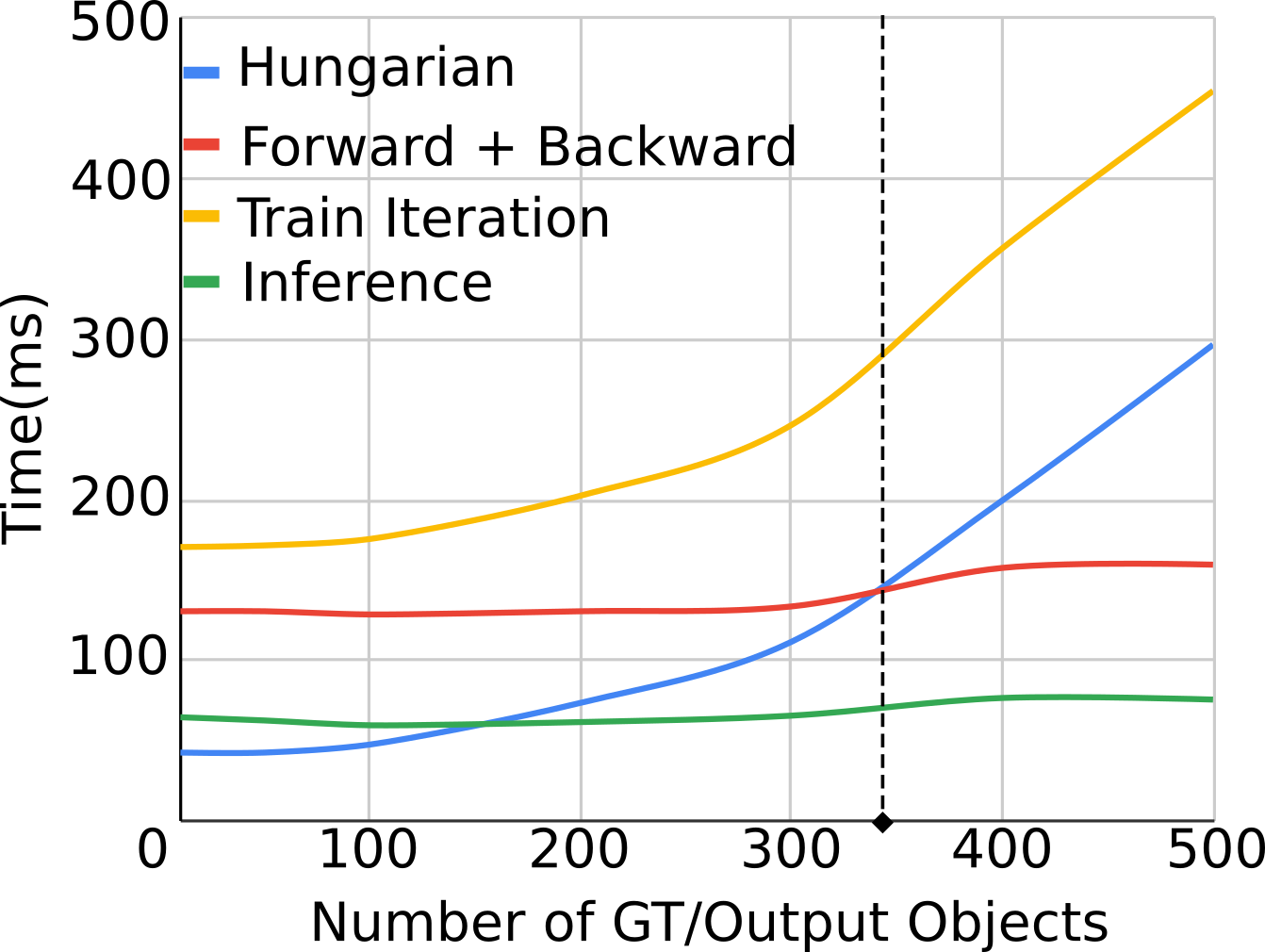}
	\caption{\revised{Computation overhead of training (yellow line) and inference (green line) iteration in millisecond clock time over number of objects. We use our set prediction model with backbone ResNet101. We assume the number of ground truth objects is fixed and is equal to the maximum number of output instances (\ie the worst case scenario). This computation overhead can increase significantly with more number of objects during training, due to the computation required for solving the assignment problem (blue line), while it does not affect inference time.}    }
	\label{fig:OverHead}
\end{figure}

\revised{Compared to the tensor-based~\cite{ren2015faster,redmon2018yolov3} object detectors, the set-based (and also the LSTM-based~\cite{Stewart:2016:CVPR}) object detectors require solving an assignment problem (\eg using Hungarian) for each input instance during each training iteration, imposing an additional computation overhead/processing time during the model training. As shown in \Fig~\ref{fig:OverHead}, the processing time for solving this assignment increases with the maximum number of objects in a dataset and it can overtake the time required for the model's weights update (a forward and a backward passes) during training, if this maximum number can reach a few hundreds. }

\revised{The results in Tab.~\ref{table:cropmot new} also indicates that a better backbone, decoder or bounding box loss can further enhance the results of our models. Note that DETR~\cite{carion2020end} can be considered as one of our suggested set-based detector baselines with a transformer-based architecture~\cite{vaswani2017attention}; However without any output to predict the cardinality. Considering F1-score as the main metric for evaluating the final predicted outputs, our best model, shown as ``{\bf Ours*}'', can outperform DETR with a similar network model and bounding box regression loss, but using the predicted cardinalities during inference. This is an expected outcome; because learning to predict a cardinality for each input data can be interpreted as an instance dependent threshold, which intuitively improves a model's performance, relying only on a single global threshold to generate its outputs.}

\begin{table*}[bt]
	\footnotesize\revised{
	\caption{\small \revised{Detection results for the set-based baselines using different backbones and decoders on (a) Pascal VOC and (b) MS COCO measured by mAP, the best F1 scores and MR. All these baselines use L1+GIoU as their bounding box regression loss. The abbreviations ``Card.'', ``Trans.'', ``En.'', and ``Dec.'' mean ``Cardinality'',``Transformer'', ``Encoder'' and ``Decoder'', respectively. } }
	\vspace{-1.em}
	\begin{center}
	\begin{tabular}{cc}
		\begin{tabular}{l|cccccc}
		\multicolumn{7}{c}{(a) PASCAL VOC}\\
			\toprule
			\shft Backbone$\slash$Decoder & Card. & mAP $\uparrow$  & F1$\uparrow$ & MR $\downarrow$ & Parameters & Flops\\
			%\cmidrule(l{2pt}r{2pt}){2-3}  \cmidrule(l{2pt}r{2pt}){4-5} 
			\midrule
			%\shft 			\revised{Faster R-CNN}              & $-$& $0.69$ & $-$ & $-$&42M&121G     \\
			\shft 			ResNet50$\slash$Trans. En.                 &$-$ & $0.62$ & $0.43$ &$0.52$& 32M & 47.8G\\
            \shft           ResNet50$\slash$Trans. Dec.                  &$-$ & $0.64$ & $0.56$  &$0.47$& 33M & 45.7G\\
            \shft           ResNet50+Trans. En.$\slash$Trans. Dec.~\cite{carion2020end}     &$-$   & $\textbf{0.70}$ & $0.71$ & $\textbf{0.42}$ & 41M & 49.4G\\
            \shft           ResNet50+Trans. En.$\slash$Trans. Dec.   &\checkmark & $-$  & $\textbf{0.72}$ &$-$ & 41M & 49.5G\\
			\bottomrule
		\end{tabular}
	&
	\begin{tabular}{ccccc}
	\multicolumn{5}{c}{(b) MS COCO}\\
			\toprule
			\shft Card.&  AP $\uparrow$  & F1$\uparrow$ & Parameters&Flops\\
			%\cmidrule(l{2pt}r{2pt}){2-3}  \cmidrule(l{2pt}r{2pt}){4-5} 
			\midrule
			%\shft 			Faster R-CNN                &$-$& $0.40$ & $-$ &42M& 180G\\
			%\shft 			YOLO V3                     &$-$& $0.33$ & $-$ &59M&141G \\
			\shft 			 $-$& $0.29$ & $0.45$&32M&83.9G\\
			\shft 			$-$ &  $0.37$ & $-$&33M&75.9G\\
			\shft 			$-$  & $0.42$ & $0.61$&41M&85.8G\\
			\shft 			\checkmark  & $-$ & $\bf{0.64}$&41M&86.0G\\
			\bottomrule
		\end{tabular}
		\end{tabular}
	\end{center}
	\label{table:voccoco}}
\end{table*}

\subsubsection{\revised{Object detection (PASCAL VOC \& COCO)}}
\begin{figure*}[t]
    \includegraphics[width=.99\linewidth]{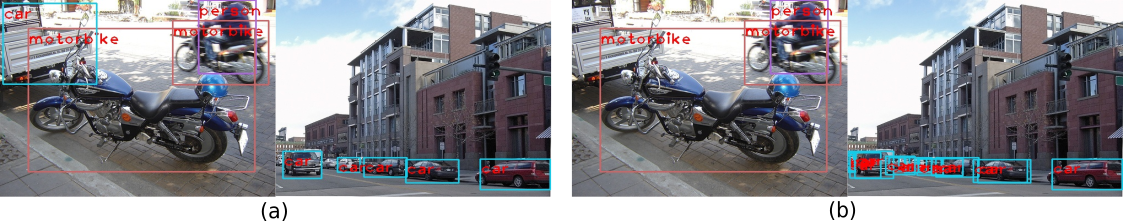}
	\small
	\caption{\revised{ A qualitative comparison for our best set-based detection baselines (ResNet50+Transformers Ecoder-decoder) with (a) the cardinality term (b) without cardinality module (\cf~\cite{carion2020end}) on PASCAL VOC detection dataset. The baseline model trained with cardinality module can effectively reduce both false positives and false negatives.}}
	\label{fig:Pascal detection}
	\vspace{-1.em}
\end{figure*}

\revised{We also evaluate the performance of our set based detection baselines on the PASCAL VOC~\cite{Everingham:2007:PASCAL-VOC} and MS COCO~\cite{lin2014microsoft} as the most popular (multi-class) object detection datasets (See \Sec~\ref{subsec:imagetagging} for their details.)}\\  
% \textbf{Dataset.} The Pascal Visual Object Classes
% (VOC) [4] benchmark is one of the most widely used
% datasets for object detection and semantic
% segmentation. It consists of 9963 images with a 50/50 split
% for training and test, where objects from 20 pre-defined
% categories have been annotated with bounding boxes.
% MS COCO: Another popular benchmark for image
% captioning, recognition, detection and segmentation is
% the more recent Microsoft Common Objects in Context
% (MS-COCO) [14].\\
\revised{\textbf{Detection Results.} Quantitative detection results for all our set-based baselines with different backbone and decoder architectures on these two datasets are shown in Tab.~\ref{table:voccoco}. We observe that our simple baseline with standard Full ResNet-101 as the backbone and the decoder fails to generate a reasonable detection performance on these two large-scale and multi-class datasets. We argue that this is mainly due to a very simple decoder, \ie few MLP layers, to output such complex unstructured data, \ie a set, from an encoded feature vector by the backbone. Compared to the pedestrian dataset, the complexities of the task, relatively larger scale images and considerably higher number of instances challenge this very simple MLP decoders for this task. To address this, we replace these MLP layers by two new, but more reliable, output layer decoders, \ie transformer encoder or decoder architecture~\cite{vaswani2017attention}. Results in Tab.~\ref{table:voccoco} verify that even with a smaller backbone model, \eg the convolutional layers of ResNet-50, in combination with a transformer encoder~\cite{vaswani2017attention} or decoder~\cite{vaswani2017attention}, our model can generate a  performance comparable to the state-of-the-art object detectors in these two datasets. The performance gap between these baselines with different backbones and decoders in MS COCO (as more challenging dataset) is more significant. This reflects the importance of a proper choice of the backbone and the decoder in these set-based detection models in challenging and large-scale datasets. }\\
\revised{Comparing with many existing tensor/anchor-based detection algorithms, DETR as a set-based detector is currently one of the best performing detectors in these two datasets with respect to mAP and AP metrics. However, our results in Tab.~\ref{table:voccoco} (last two rows) indicate that our full model with the cardinality module can improve the performance our DETR baseline with respect to F1 score. \Fig~\ref{fig:Pascal detection} represents few detection examples comparing DETR model detection output using the best threshold against our model using the cardinality value during the inference. }\\

% We can also conclude that     model compared to the pedestrian dataset result in failure to  the image sizes and the maximum number of objects was considerably smaller than those in PASCAL VOC and COCO datasets. the    We have found the pure resnet structurehas certain limits when it comes to set prediction with scale. Firstly, its last layer is a simple MLPwhich struggles to carry through objects information at large scale.    \\

% \begin{table*}[bt]
% 	\footnotesize\revised{
% 	\caption{\small \revised{Detection results of MS COCO measured by AP, the best F1 scores.} }
% 	\begin{center}
% 		\begin{tabular}{lccccc}
% 			\toprule
% 			\shft Backbone$\slash$Decoder &Card.&  AP $\uparrow$  & F1$\uparrow$ & Parameters&Flops\\
% 			%\cmidrule(l{2pt}r{2pt}){2-3}  \cmidrule(l{2pt}r{2pt}){4-5} 
% 			\midrule
% 			%\shft 			Faster R-CNN                &$-$& $0.40$ & $-$ &42M& 180G\\
% 			%\shft 			YOLO V3                     &$-$& $0.33$ & $-$ &59M&141G \\
% 			\shft 			ResNet50$\slash$Trans. En.                 &$-$& $0.29$ & $0.45$&32M&83.9G\\
% 			\shft 			ResNet50$\slash$Trans. Dec.                  &$-$ &  $0.37$ & $-$&33M&75.9G\\
% 			\shft 			ResNet50+Trans. En.$\slash$Trans. Dec.     &$-$  & $0.42$ & $0.58$&41M&85.8G\\
% 			\shft 			ResNet50+Trans. En.$\slash$Trans. Dec.     &\checkmark  & $-$ & $\bf{0.63}$&41M&86.0G\\
% 			\bottomrule
% 		\end{tabular}
% 	\end{center}
% 	\label{table:cropmot}}
% \end{table*}

\subsection{CAPTCHA test for de-summing a digit (Scenario 3)}
\begin{figure}
	%\vspace{-0.8em}
	\includegraphics[width=0.99\linewidth]{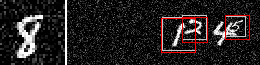}
	\small
	\vspace{-0.8em}
	\caption{\small A \emph{query digit} (left)  and a \emph{set of digits} (right) for the proposed CAPTCHA test. The ground truth and our predicted solutions are shown by white and red boxes respectively. }
	\label{fig:captcha_test}
	\vspace{-1.0em}
\end{figure}

We also evaluate our set formulation on a CAPTCHA test where the aim is to determine whether a user is a human or not by a moderately complex logical test. In this test, the user is asked to decompose a \emph{query digit} shown as an image (Fig.~\ref{fig:captcha_test} (left)) into a set of digits by clicking on a subset of numbers in a noisy image (Fig.~\ref{fig:captcha_test} (right)) such that the summation of the selected numbers is equal to the \emph{query digit}. 

In this puzzle, it is assumed there exists only one valid solution (including an empty response). We target this complex puzzle with our set learning approach. What is assumed to be available as the training data is a set of spotted locations in the \emph{set of digits} image and no further information about the represented values of \emph{query digit} and the \emph{set of digits} is provided. In practice, the annotation can be acquired from the users' click when the test is successful. In our case, we generate a dataset for this test from the real handwriting MNIST dataset. 
%\fixmeh{Farbod, can you write some details here how you generated the data from MNIST. Not too much details though}

\textbf{Data generation.}
The dataset is generated using the MNIST dataset of handwritten digits.
 The \emph{query digit} is generated by randomly selecting one of the digits from MNIST dataset. Given a \emph{query digit}, we create a series of random digits with different length such that there exists a subset of these digits that sums up to the \emph{query digit}. 
Note that in each instance there is only one solution (including empty) to the puzzle.
We place the chosen digits in a random position on a $300\times75$ blank image with different rotations and sizes. To make the problem more challenging, random white noise is added to both the \emph{query digit} and the \emph{set of digits} images (Fig.~\ref{fig:captcha_test}). The produced dataset includes $100$K problem instances for training and $10$K images for evaluation, generated independently from MNIST training and test sets.  

\textbf{Baseline methods.}
%\fixmeh{Farbod, can you write some details here, I will polish later}
Considering the fact that only a set of locations is provided as ground truth, this problem can be seen as an analogy to the object detection problem. However, the key difference between this logical test and the object detection problem is that the objects of interest (the selected numbers) change as the \emph{query digit} changes. For example, in Fig.~\ref{fig:captcha_test}, if the \emph{query digit} changes to another number, \eg $4$, the number $\{4\}$ should be only chosen. Thus, for the same \emph{set of digits}, now $\{1,2,5\}$ would be labeled as background. Since any number can be either background or foreground conditioned on the \emph{query digit}, this problem cannot be trivially formulated as an object detection task. To prove this claim, as a baseline, we attempt to solve the CAPTCHA problem using a detector, \eg \revised{Faster R-CNN}, with the same base structure as our network (ResNet-101) and trained on the exactly same data including the \emph{query digit} and the \emph{set of digits} images\footnote{To ensure inputting one single image into the ResNet-101 backbone, we represent both the \emph{query digit} and \emph{set of digits} by a single image such that the \emph{query digit} always appears in the bottom left corner of the \emph{set of digits} image.}.  
%\fixmea{What is the input? How to you provide the query image at training and test time? Both to R-CNN and our network?}

As an additional baseline, we also present a solution to this set problem using the LSTM-based detection model~\cite{vinyals2015order, Stewart:2016:CVPR}, similar to the framework used in \Sec~\ref{result:od}. To this end, we feed the inputs to the same base structure as our network (ResNet-101) and replace the last fully connected layer with a one layer LSTM, which is used to predict in each iteration the parameters of a bounding box, \ie $\by = (x_1,y_1,x_2,y_2,s)$. During training, we use Hungarian algorithm to match the predicted bounding boxes with the ground truths. During inference, the predicted bounding boxes with a score above $0.5$ are represented as the final outputs. 
%As an additional baseline, we also try to solve this problem using an additional classifier in order to recognize the query number and the set of digits detected by the Faster R-CNN in the previous step. Since we assume that no information about the represented values of query digit and the set of digits are provided, the classifier (ResNet-101) is trained on MNIST dataset. Then, we try to solve this problem by recognizing the digits. 

\textbf{Implementation details.}
%\fixmeh{Farbod, can you write some details here, I will polish later. Check my points for Roman}
We use the same set formulation as in the previous experiment on object detection.
Similarly, we train the same network structure (ResNet-101) using the same optimizer and hyper-parameters as described in \ref{result:od}. Similar to the previous experiment, \revised{we do not use} the permutation loss $\calL_{perm}(\cdot)$ (\textbf{Scenario 2}) since we are not interested in the permutation distribution of the detected digits in this experiment. However, we still need to estimate the permutations iteratively using Eq.~\ref{eq:alternation1b} to permute the ground truth for $\calL_{state}(\cdot)$. 

The input to the network is both the \emph{query digit} and the \emph{set of digits} images and the network outputs bounding boxes corresponding to the solution set. The hyper-parameter $U$ is set to be $2$, adjusted on the
validation set.

\textbf{Evaluation protocol.}
Localizing the numbers that sum up to the query digit is important for this task, therefore, we evaluate the performance of the network by comparing the ground truth with the predicted bounding boxes. More precisely, to represent the degree of match between the prediction and ground truth, we employ the commonly used Jaccard similarity coefficient. If $IoU_{(b1, b2)} > 0.5$ for all the numbers in the solution set, we mark the instance as correct otherwise the problem instance is counted as incorrect. 
%\fixmea{I assume for the solution to be correct, none of the numbers not in the solution set should have an IOU > 0.5 with any predicted box.}

\textbf{Results.}
\begin{table}
	\small
	\caption{\small Accuracy for solving the CAPTCHA test.}
	\begin{center}
		\begin{tabular}{ccc}
			\toprule
			\revised{Faster R-CNN} & LSTM+Hungarian & \bf Ours  \\
			%\cmidrule(l{2pt}r{2pt}){2-3}  \cmidrule(l{2pt}r{2pt}){4-5} 
			\midrule
			\shft 			$26.8\%$ & $66.1\%$ & $\bf 95.6\%$\\
			\bottomrule
		\end{tabular}
	\end{center}
	\label{table:captcha}
	\vspace{-1.7em}
\end{table}\
The accuracy for solving this test using all competing methods is reported in \Tab~\ref{table:captcha}. As expected, \revised{Faster R-CNN} fails to solve this test and achieves an accuracy of 26.8\%. This is because \revised{Faster R-CNN} only learns to localize digits in the image and ignores the logical relationship between the objects of interest. A detection framework is not capable of performing reasoning in order to generate a sensible score for a subset of objects (digits). The LSTM + Hungarian network is able to approximate the mapping between inputs and the outputs as it can be trained in an end-to-end fashion. However, it is worse than our approach at solving this problem, mainly due to the assumption about sequential dependency between the outputs (even with an unknown ordering/permutation), which is not always satisfied in this experiment.  In contrast, our set prediction formulation provides an accurate model for this problem and gives the network the ability of mimicking arithmetic implicitly by end-to-end learning the relationship between the inputs and outputs from the training data. In fact, the set network is able to generate different sets with different states and cardinality if one or both of the inputs change. This shows the potential of our formulation to tackle arithmetical, logical or semantic relationship problems between inputs and output sets without any explicit knowledge about arithmetic, logic or semantics.

\section{Conclusion}
In this paper, we proposed a framework for predicting
sets with unknown cardinality and permutation using convolutional neural networks. In our formulation, set permutation is considered as an unobservable variable and its distribution is estimated iteratively using alternating optimization. We have shown that object detection can be elegantly formulated as a set prediction problem, where a deep network can be learned end-to-end to generate the detection outputs with no heuristic involved. We have demonstrated that the approach is able to outperform the state-of-the art object detections on real data including highly occluded objects. 
%
% In addition, the proposed framework can simultaneously detect and identify the similarly looking object instances using the set permutation, which is applicable for applications such as multiple object tracking. 
%
We have also shown the effectiveness of our set learning approach on solving a complex logical CAPTCHA test, where the aim is to de-sum a digit into its components by selecting a set of digits with an equal sum value. 

The main limitation of the current
framework for {\bf Scenario. 2} is that the number of possible permutations exponentially grows with the maximum set size (cardinality). Therefore, applying it to large-scale problem is not straightforward and requires an accurate approximation for estimating a subset of dominant permutations. In future, we plan to overcome this limitation by learning the subset of significant permutations to target real-world large-scale problems such as multiple object tracking.

\noindent{\bf Acknowledgements.} Ian Reid and Hamid Rezatofighi gratefully acknowledge the support the Australian Research Council through Laureate Fellowship FL130100102 and Centre of Excellence for Robotic Vision CE140100016. This work, also, was partially funded by the Sofja Kovalevskaja Award from the Humboldt Foundation.
% or


% \appendices

% % you can choose not to have a title for an appendix
% % if you want by leaving the argument blank
% \section{}

% % use section* for acknowledgment
% \ifCLASSOPTIONcompsoc
%   % The Computer Society usually uses the plural form
%   \section*{Acknowledgments}
% \else
%   % regular IEEE prefers the singular form
% \fi


\ifCLASSOPTIONcaptionsoff
  \newpage
\fi



\bibliographystyle{IEEEtran}
\bibliography{ref,refs-short,anton-ref}

% \begin{IEEEbiography}[{\includegraphics[width=1in,height=1.25in,clip,keepaspectratio]{mshell}}]{Tianyu Zhu}


%\vfill

\begin{IEEEbiography}
    [{\includegraphics[width=1in,height=1.25in,clip,keepaspectratio]{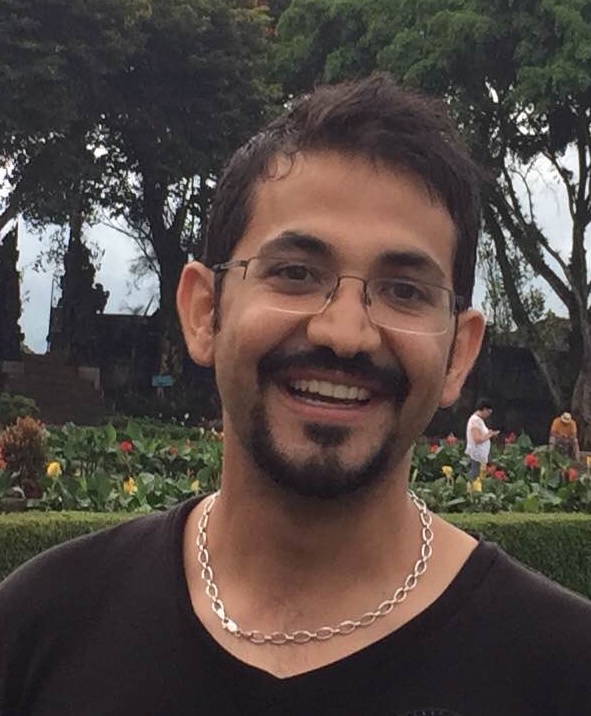}}]{Hamid Rezatofighi}
is a lecturer at the Faculty of
Information Technology, Monash University, Australia. Before that, he was an Endeavour Research
Fellow at the Stanford Vision Lab (SVL), Stanford
University and a Senior Research Fellow at the
Australian Institute for Machine Learning (AIML),
The University of Adelaide. His main research interest focuses on computer vision and vision-based
perception for robotics, including object detection,
multi-object tracking, human trajectory and pose
forecasting and human collective activity recognition. He has also research expertise in Bayesian filtering, estimation and
learning using point process and finite set statistics
\end{IEEEbiography}

\vspace{-2em}
\begin{IEEEbiography}
    [{\includegraphics[width=1in,height=1.25in,clip,keepaspectratio]{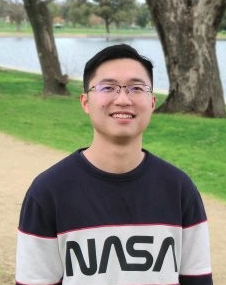}}]{Alan Tianyu Zhu}
is a computer vision PhD candidate at Monash University since 2018. He has received Bachelor of Engineering and Science from Monash University with first class Honours in 2017. He is also a machine learning research associate at Faculty of IT, Monash University on learning with less labels projects. His current research focus is attention mechanisms, object detection and multi object tracking. 
\end{IEEEbiography}

\vspace{-2em}
\begin{IEEEbiography}
    [{\includegraphics[width=1in,height=1.25in,clip,keepaspectratio]{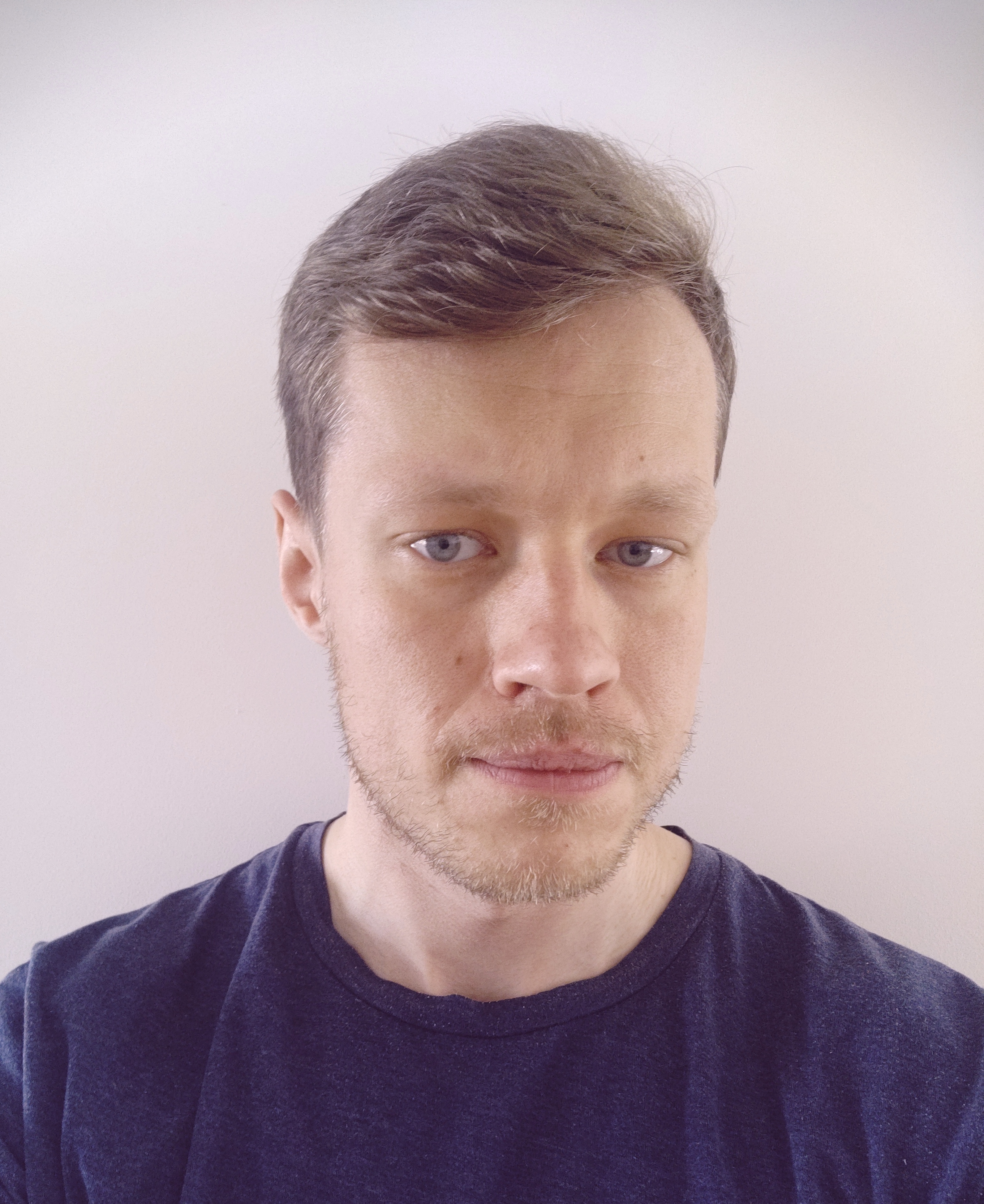}}]{Roman Kaskman} received his B.Sc. degree in Computer Science from Tallinn University of Technology, Estonia, in 2015 and M.Sc. degree in Computer Science from Technical University of Munich, Germany, in 2019. He was a student research assistant at Siemens Corporate Technology from 2018 to 2020. Since 2020 he is working at SnkeOS GmbH as a Machine Learning Engineer.

\end{IEEEbiography}
\vspace{-2em}
\begin{IEEEbiography}
    [{\includegraphics[width=1in,height=1.25in,clip,keepaspectratio]{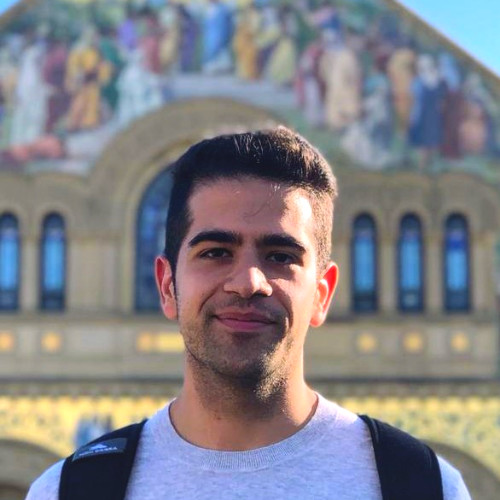}}]{Farbod T. Motlagh} is a Software Engineer at Google. He studied Computer Science at the University of Adelaide and graduated with first class Honours in 2019. During his honours degree, he was a research assistant at Australian Institute for Machine Learning (AIML) where he worked on unsupervised representation learning and metric learning.

\end{IEEEbiography}

\vspace{-2em}
\begin{IEEEbiography}
    [{\includegraphics[width=1in,height=1.25in,clip,keepaspectratio]{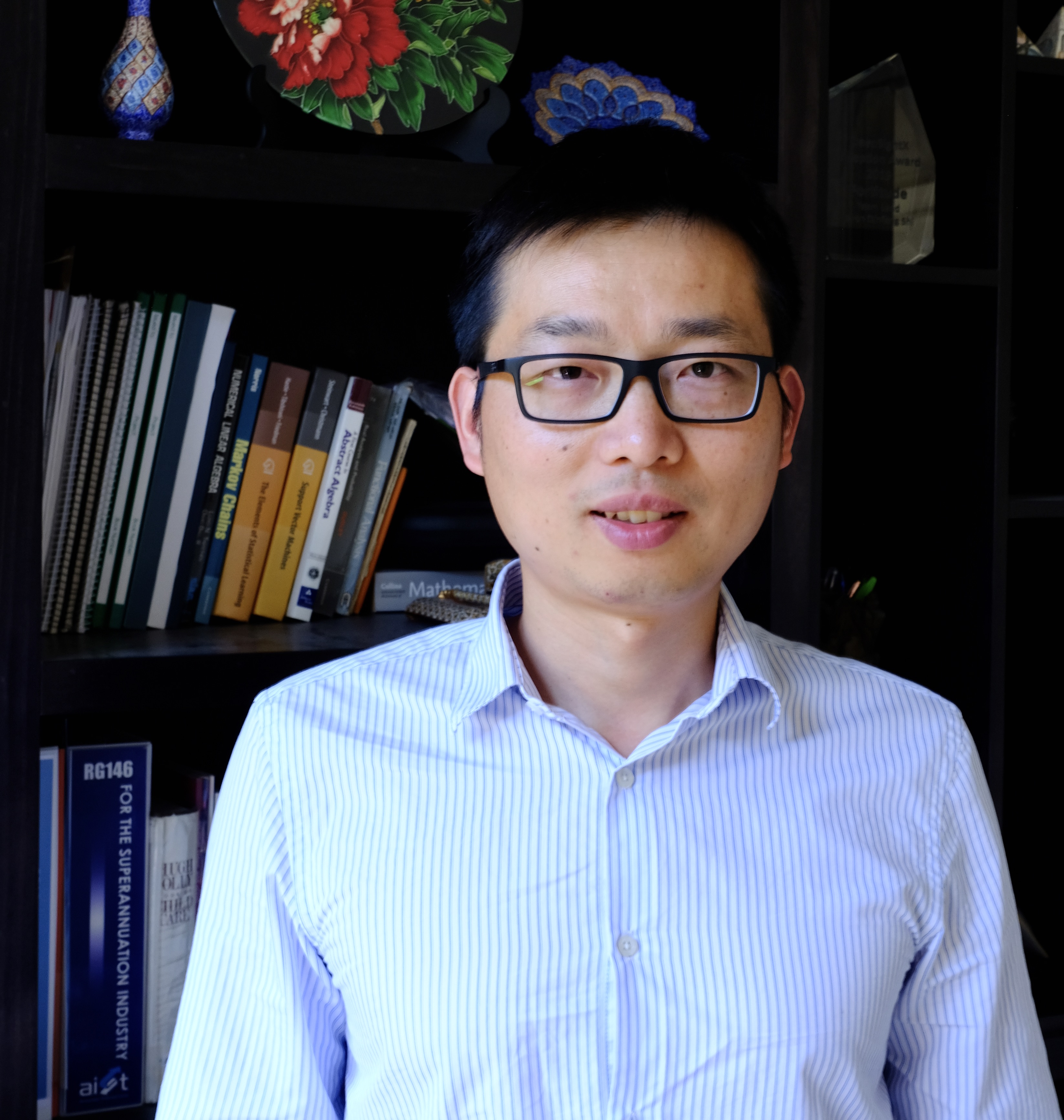}}]{Javen Qinfeng Shi}
is the Founding Director of Probabilistic Graphical Model Group at the University of Adelaide, Director in Advanced Releasing and Learning of Australian Institute for Machine Learning (AIML), and the Chief Scientist of Smarter Regions. He is a leader in machine learning in both high-end AI research and also real world applications with high impacts.
He is recognised both locally and internationally for the impact of his work, through an impressive record of publishing in the highest ranked venue in the field of Computer Vision \& Pattern Recognition (22 CVPR papers), through an impeccable record of ARC funding (including his DECRA, 2 DP as 1st CI, 4 LP as Co-CI). 

\end{IEEEbiography}

\vspace{-2em}
\begin{IEEEbiography}
    [{\includegraphics[width=1in,height=1.25in,clip,keepaspectratio]{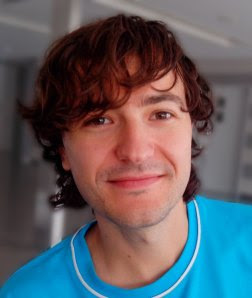}}]{Anton Milan}
is a Sr. Applied Scientist at Amazon Robotics AI. He studied Computer Science and Philosophy at the University of Bonn and received his PhD in May 2013. In 2014 he joined the Australian Centre for Visual Technologies (ACVT) at the University of Adelaide as a post-doctoral researcher. He was at ETH in 2015 and at the University of Bonn in 2016 as a visiting researcher. 
\end{IEEEbiography}

\vspace{-2em}
\begin{IEEEbiography}
    [{\includegraphics[width=1in,height=1.25in,clip,keepaspectratio]{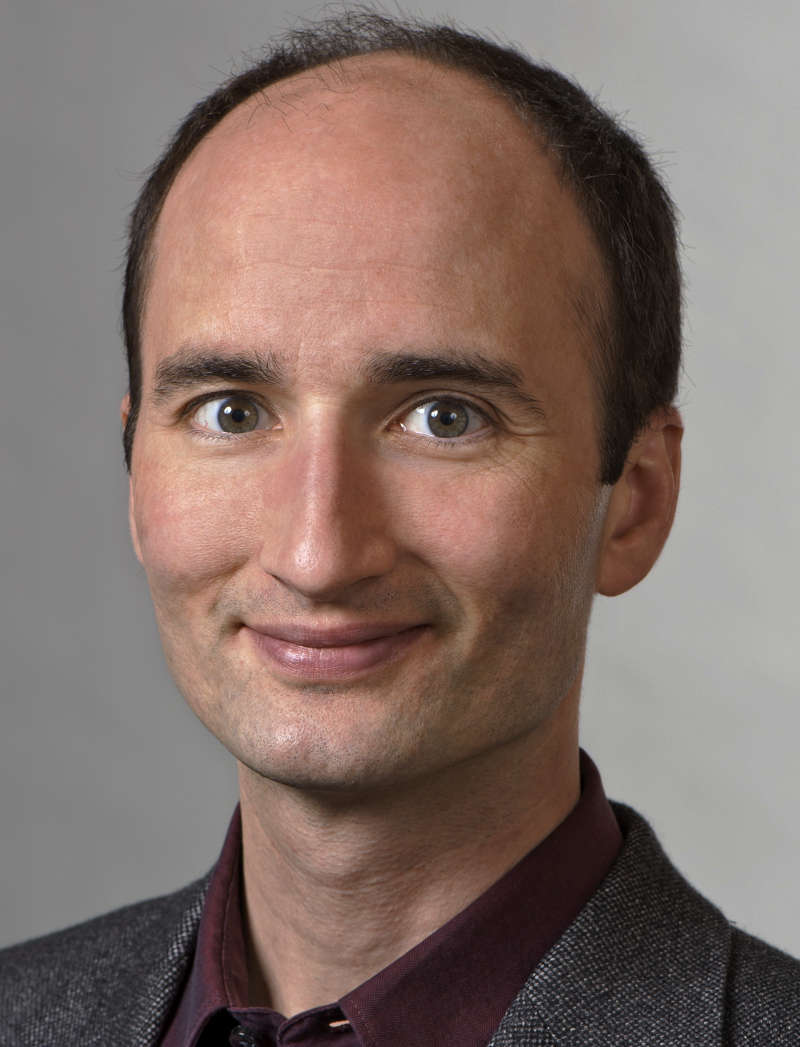}}]{Daniel Cremers}
studied physics and mathematics in Heidelberg and New
York. After his PhD in Computer Science he spent two years at UCLA and
one year at Siemens Corporate Research in Princeton. From 2005 until
2009 he was associate professor at the University of Bonn. Since 2009
he holds the Chair of Computer Vision and Artificial Intelligence at
the Technical University of Munich.  In 2018
he organized the largest ever European Conference on Computer Vision
in Munich. He is member of the Bavarian Academy of Sciences and
Humanities.  In 2016 he received the Gottfried Wilhelm Leibniz Award,
the biggest award in German academia. He is co-founder and chief
scientific officer of the high-tech startup Artisense.
\end{IEEEbiography}

\vspace{-2em}
\begin{IEEEbiography}
    [{\includegraphics[width=1in,height=1.25in,clip,keepaspectratio]{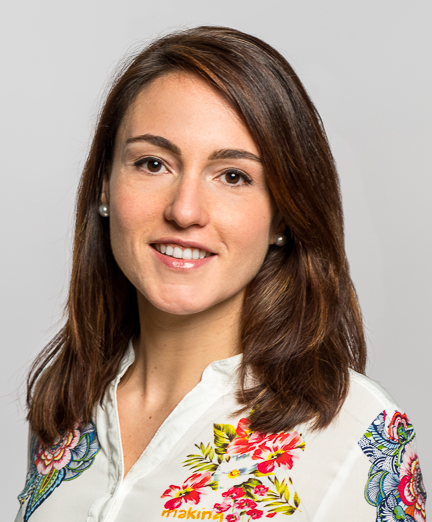}}]{Laura Leal-Taix{\'e}}
is s Professor at the Technical University
of Munich, Germany. She received her Bachelor
and Master degrees in Telecommunications Engineering from the Technical University of Catalonia (UPC), Barcelona. She did her Master
Thesis at Northeastern University, Boston, USA
and received her PhD degree (Dr.-Ing.) from the
Leibniz University Hannover, Germany. During
her PhD she did a one-year visit at the Vision
Lab at the University of Michigan, USA. She
also spent two years as a postdoc at the Institute of Geodesy and
Photogrammetry of ETH Zurich, Switzerland and one year at the Technical University of Munich. She is the recipient of the prestigious Sofja Kovalevskaja Award from the Humboldt Foundation and the Google Faculty Award. Her research interests are dynamic scene
understanding, in particular multiple object tracking and segmentation,
as well as machine learning for video analysis.
\end{IEEEbiography}

\vspace{-2em}
\begin{IEEEbiography}
    [{\includegraphics[width=1in,height=1.25in,clip,keepaspectratio]{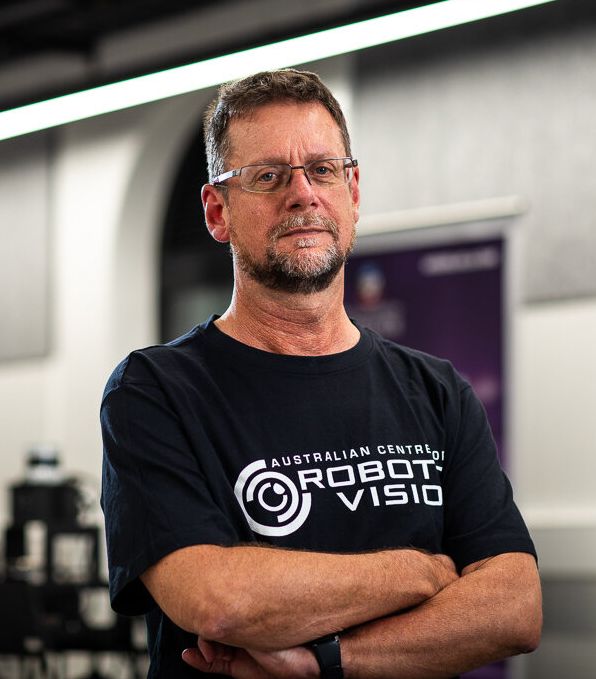}}]{Ian Reid } is the Head of the School of Computer Science at the University of Adelaide. He is a Fellow of Australian Academy of Science and Fellow of the Academy of Technological Sciences and Engineering, and held an ARC Australian Laureate Fellowship 2013-18. Between 2000 and 2012 he was a Professor of Engineering Science at the University of Oxford.  His research interests include robotic vision, SLAM, visual scene understanding and human motion analysis. He has published 330 papers. He serves on the program committees of various national and international conferences, including as Program Chair for the Asian Conference on Computer Vision 2014 and General Chair Asian Conference on Computer Vision 2018. He was an Area Editor for T-PAMI, 2010-2017.
\end{IEEEbiography}


%\input{appendix}

\end{document}